\documentclass{article}

\usepackage[final]{corl_2026}
\makeatletter
\renewcommand{\@conferencelocation}{Austin, Texas, USA}
\makeatother
\usepackage{amsmath}
\usepackage{amssymb}
\usepackage{graphicx}
\usepackage[percent]{overpic}
\usepackage{subcaption}
\usepackage{booktabs}
\usepackage{wrapfig}
\usepackage{xcolor}
\usepackage{colortbl}
\usepackage{multirow}
\usepackage[most]{tcolorbox}
\usepackage{afterpage}

\title{The Speedup Paradox: Rethinking Inference Speed-Quality Trade-off in Embodied Tasks}

\author{
\textbf{Yujin Wang \quad Junli Chen$^{\dagger}$ \quad Yixuan Li$^{\dagger}$ \quad Shunan Dong}\\
\textbf{Huazhong Yang \quad Yongpan Liu \quad Hongyang Jia}\\
Tsinghua University\\
{\small\texttt{\char123 yujin-wa24,jl-chen22,liyixuan23\char125 @mails.tsinghua.edu.cn}\quad
\texttt{hjia@tsinghua.edu.cn}}\\
{\small \quad $^{\dagger}$ equal contribution.}
}

\hypersetup{
    pdftitle={The Speedup Paradox: Rethinking Inference Optimization in Embodied Tasks},
    pdfauthor={Yujin Wang, Junli Chen, Yixuan Li, Shunan Dong, Huazhong Yang, Yongpan Liu, Hongyang Jia}
}

\begin{document}
\maketitle

\addtocontents{toc}{\protect\setcounter{tocdepth}{-1}}

\begingroup
\setlength{\intextsep}{6pt plus 1pt minus 1pt}
\vspace{-1em}
\begin{figure*}[h!]
    \centering
    \setlength{\abovecaptionskip}{2pt}
    \includegraphics[width=0.92\textwidth]{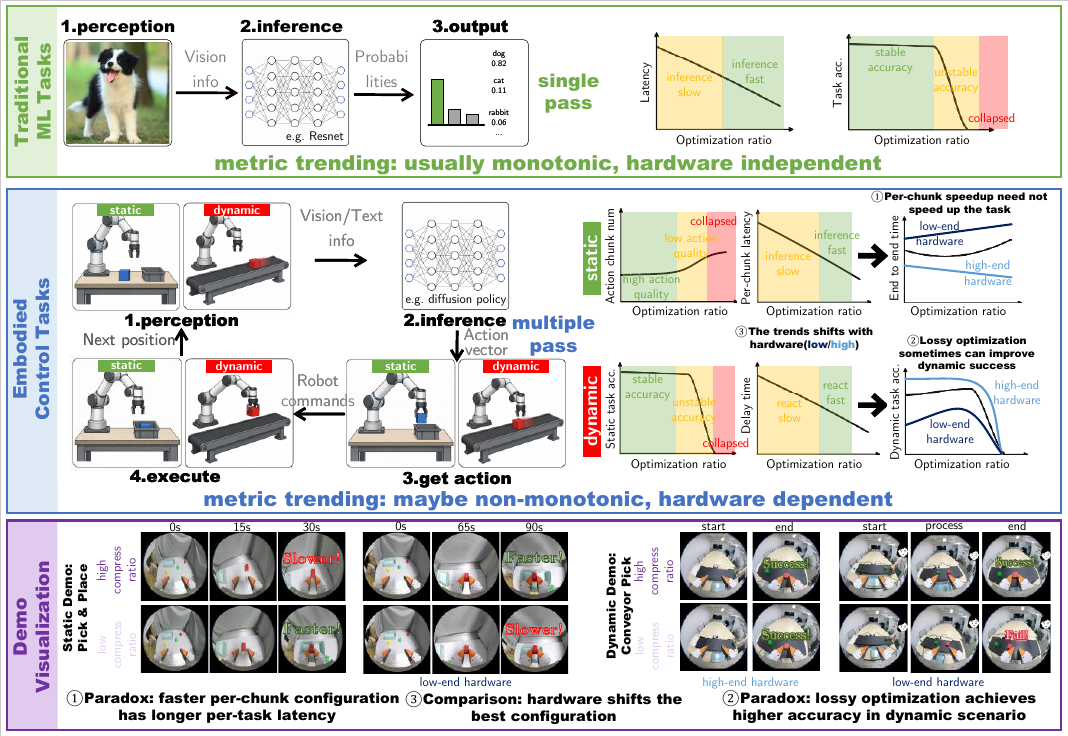}
    \caption{Overview of the speed-quality trade-off in embodied inference.
    Unlike static ML tasks, where stronger lossy lightweighting optimization monotonically trades accuracy for inference speed,
    lightweighting embodied policies yields non-monotonic task-level behavior: both per-task time (static tasks) and task success rate (dynamic tasks) may exhibit a sweet spot,
    whose location further shifts with the deployment hardware.}
    \label{fig:overview}
    \vspace{-1.8em}
\end{figure*}
\endgroup

\begin{abstract}
Embodied foundation models have recently been widely used to improve robot generalization and task success rates.
Previous works apply lossy efficient-inference techniques such as quantization, pruning, and asynchronous inference,
accepting small action quality degradation in exchange for lower per-inference computation cost and inter-action latency.
However, unlike traditional static ML tasks, embodied tasks involve repeated interaction with the environment,
and task-level performance is determined not only by inference cost,
but also by closed-loop effects unique to embodied execution,
which remain insufficiently characterized in current efficient-inference studies.
In this work, we propose TISED (\underline{T}ask-level \underline{I}nference \underline{S}peedup \underline{E}ffect \underline{D}ecomposition),
an analytical framework that unifies diverse lossy lightweighting inference optimization techniques and decomposes their effects on static and dynamic tasks;
using it, we uncover several paradoxical effects:
(1) on \textit{static tasks}, optimization can sometimes  lengthen end-to-end per-task completion time even as per-chunk time drops;
(2) on \textit{dynamic tasks}, moderate lossy lightweighting optimization can raise task success rate even above the baseline; and
(3) the monotonicity and sweet-spot location of both effects can shift with hardware configuration.
Together, our findings provide a new perspective on adapting inference optimization techniques to embodied tasks.
Source code: \url{https://github.com/thu-ee-acts-lab/robotics-the-speedup-paradox}.
\end{abstract}


\section{Introduction}

Embodied models such as Vision-Language-Action models (VLA)~\cite{kim2024openvla, shukor2025smolvla, black2024pi0, black2025pi0.5, brohan2023rt1roboticstransformerrealworld, rt22023arxiv, octo_2023, liu2024rdt} and
World-Action Models (WAM)~\cite{kim2026cosmos, li2026causal, ye2026worldactionmodelszeroshot, ye2026gigaworldpolicyefficientactioncenteredworldaction}
have been widely deployed across diverse embodied scenarios.
These models are typically obtained by fine-tuning large-scale pretrained vision-language models~\cite{zhai2023sigmoid, beyer2024paligemma} and world models~\cite{agarwal2025cosmos, hafner2023mastering},
yielding strong cross-task generalization and improved robustness to environmental variation.
However, to fully accommodate these capabilities, they usually have large parameter counts and compute FLOPs at inference time,
posing a key barrier to edge deployment and real-time control.

To address this, many recent works apply lossy lightweighting inference optimization techniques to embodied models, such as
quantization~\cite{zhang2026quantvlascalecalibratedposttrainingquantization, xu2026qvlachannelsequalvisionlanguageaction},
cache reuse~\cite{xu2025vla, li2025sp, wang2025specprune}, sampling step reduction~\cite{lu2022dpm, chi2024diffusionpolicyvisuomotorpolicy, wang2024onestepdiffusionpolicyfast}, and asynchronous scheduling~\cite{tang2025vlash, shi2026streamingvla, lu2026fasterrethinkingrealtimeflow}.
These methods have been shown to reduce inference latency or shorten the gap between consecutive action chunks with slight accuracy degradation.
Embodied tasks, however, differ fundamentally:
Each chunk of the inference-execution loop consists of one policy inference and the subsequent action execution, and the loop runs for many such chunks before a rollout terminates.
The reductions in inference latency from these methods therefore capture only a fragment of task-level performance:
they ignore how many chunks the rollout ultimately requires until success,
how the environment evolves during model inference,
and how all of these interact with hardware performance across inference and execution costs.

To analyze how lossy lightweighting inference optimization shapes embodied task performance beyond inference latency,
we propose TISED (\underline{T}ask-level \underline{I}nference \underline{S}peedup \underline{E}ffect \underline{D}ecomposition) analysis shown in Figure~\ref{fig:overview}.
First, we place policy-intrinsic and execution-aware optimization on a common per-chunk time formula.
Building on this unified view, we model the end-to-end objective separately for the two task regimes: on static tasks, end-to-end
completion time is jointly determined by the per-chunk time and the number of chunks that the full rollout requires; on dynamic
tasks, task success rate is jointly determined by action quality and
observation staleness.

We empirically test these predictions with a progressive-lightweighting sweep across task types, model backbones,
optimization methods, and compute platforms, from which three observations emerge:
\textbf{(1)} On static tasks, per-task time is sometimes
\emph{non-monotonic} in the lightweighting strength due to the deterioration of action quality
(inference latency reduction $\neq$ end-to-end per-task time reduction).
\textbf{(2)} On dynamic tasks, task success rate is sometimes
\emph{non-monotonic} in the lossy lightweighting optimization strength
due to the trade-off between degraded action quality and reduced observation staleness
(static task accuracy $\neq$ dynamic task accuracy).
\textbf{(3)} With the task and lightweighting method fixed, the above trends and the
location of the sweet spot shift with the
hardware compute budget.
Together, these three observations show that the inference speed-quality trade-off familiar from traditional static ML no longer predicts task-level outcomes in embodied closed-loop execution—speed and quality interact through rollout length, action delay, and hardware budget.

\vspace{-6pt}

\section{Related Works}
\vspace{-6pt}
\subsection{Embodied Foundation Models}
\vspace{-6pt}
Recent progress in vision-language-action models (VLA)~\cite{kim2024openvla, shukor2025smolvla, black2024pi0, black2025pi0.5, brohan2023rt1roboticstransformerrealworld, rt22023arxiv, octo_2023, liu2024rdt} and world-action models (WAM)~\cite{kim2026cosmos, li2026causal, ye2026worldactionmodelszeroshot, ye2026gigaworldpolicyefficientactioncenteredworldaction}
has enabled stronger open-world generalization, task adaptation, and large-scale policy transfer.
However, these capabilities come at a high inference cost. For example, by default $\pi_0$~\cite{black2024pi0} (3.3B parameters)
is deployed on high-end consumer GPUs such as RTX~4090, and Cosmos-Policy~\cite{kim2026cosmos} (2B parameters) on datacenter-class accelerators such as H100.
This is a key barrier to edge deployment and low-latency control.
In this work, rather than scaling capabilities or enhancing generalization, we study how to deploy these models efficiently in the inference-execution loop.

\vspace{-6pt}
\subsection{Lossy Lightweighting Optimizations for Embodied Model Inference}
\vspace{-3pt}
To accelerate the inference of embodied models,
recent work has explored a range of lossy lightweighting inference strategies.
These methods can be grouped into two categories:
(1) \emph{Policy-intrinsic optimization}, which reduces the computational cost of policy inference through techniques such as quantization~\cite{zhang2026quantvlascalecalibratedposttrainingquantization, xu2026qvlachannelsequalvisionlanguageaction, bitvla, yan2026hbvlapushing1bitposttraining}, denoise step reduction~\cite{ji2026sparseactiongenacceleratingdiffusion, prasad2024consistencypolicyacceleratedvisuomotor, clemente2025twostepsdiffusionpolicyrobotic, li2026onestepflowpolicyselfdistillation}, and cache reuse~\cite{xu2025vla, li2025sp, wang2025specprune, zhou2026characterizingvisionlanguageactionmodelsxpus}.
For example, Q-VLA~\cite{xu2026qvlachannelsequalvisionlanguageaction} analyzes how different components of a VLA vary in quantization sensitivity and assigns mixed precision accordingly to minimize the resulting action error;
VLA-Cache~\cite{xu2025vla} exploits the sparsity of background visual information across perception steps and reuses less-salient tokens to reduce redundant computation.
(2) \emph{Execution-aware optimization}, which improves effective control frequency by overlapping inference with execution, such as asynchronous scheduling~\cite{tang2025vlash,shi2026streamingvla,lu2026fasterrethinkingrealtimeflow}.
For example, VLASH~\cite{tang2025vlash} rolls the robot state forward using the previously generated action chunk to predict the state at execution time,
and further reduces prediction-execution misalignment by fine-tuning.
Most of these works focus only on reducing the latency of a single inference, leaving a systematic analysis of end-to-end effects in embodied scenarios largely unexplored.
In this work, rather than targeting a specific lightweight algorithm, we provide a unified modeling framework to analyze the impact of such algorithms on embodied tasks.

\vspace{-6pt}
\subsection{Efficiency-Accuracy Trade-off Characterizations}
\vspace{-3pt}
Aggressive lossy lightweighting optimization methods can achieve significant reductions in latency or energy efficiency while reducing accuracy.
Previous works such as Deep Compression~\cite{han2015deep} established a trade-off between efficiency and accuracy in deep learning models.
Subsequently, a lot of research has attempted to push this Pareto frontier forward in vision and language models~\cite{howard2017mobilenetsefficientconvolutionalneural, xiao2023smoothquant, lin2023awq,
  liang2021pruningquantizationdeepneural, zhu2024surveymodelcompressionlarge}.
In embodied systems, this relationship becomes more complex due to error accumulation and closed-loop feedback.
Works such as~\cite{Li2020StreamingP, kang2026win} introduced a new trade-off between efficiency and accuracy in time-sensitive tasks, but their scope was limited to object detection or LLM agents.
Works such as~\cite{zhou2026characterizingvisionlanguageactionmodelsxpus, jiang2026fastirunvla, taherin2026crossplatformscalingvisionlanguageactionmodels} have characterized the inference performance of VLA models across various hardware devices but do not focus on task-level performance.
Work~\cite{li2026inference} provides valuable analysis of the negative impact of model lightweighting on execution in embodied task scenarios, while their focus is on the negative side.
In this work, we further explore the interplay between the positive and negative effects of lightweight optimization and conduct a cross-hardware study.

\begingroup
\setlength{\abovedisplayskip}{4pt plus 1pt minus 1pt}
\setlength{\belowdisplayskip}{3pt plus 1pt minus 1pt}
\setlength{\abovedisplayshortskip}{0pt plus 1pt}
\setlength{\belowdisplayshortskip}{3pt plus 1pt minus 1pt}
\vspace{-10pt}
\section{Task-Level Inference Speedup Effect Decomposition}
\label{sec:unified-view}
\vspace{-8pt}
Embodied inference optimization operates within an inference-execution loop instead of during a single forward pass.
Therefore, speeding up an individual inference-execution chunk does not necessarily translate into a task-level gain: the optimization can change both how many control chunks a rollout needs and how stale the observation is when the resulting action is executed.
These two effects can give rise to different failure modes in different task regimes.
We develop this view into TISED, which analyzes task-level performance across lightweighting strength, task dynamics, and hardware budget.
TISED places policy-intrinsic and execution-aware optimization on a common closed-loop time axis, and separates the end-to-end effect into chunk time, rollout length, and action delay.

We use two task regimes throughout the paper.
In a \emph{static task}, the task-relevant environment state remains effectively unchanged during model inference (e.g., static pick \& place task).
In this regime, lightweighting can affect movement smoothness and total execution time.
In a \emph{dynamic task}, the task-relevant environment state evolves while the model is still computing the next action (e.g., picking a rolling ball or playing ping-pong).
In this regime, lightweighting can affect both action quality and reaction speed -- the inference speed can affect the policy's ability to react to dynamic environments.
A more detailed analysis and derivation can be found in Appendix~\ref{app:analysis}.

\afterpage{%
\begin{figure}[t]
    \centering
    \includegraphics[width=\linewidth]{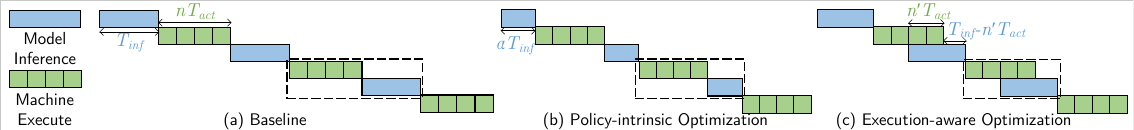}
    \setlength{\abovecaptionskip}{4pt}
    \caption{Dataflow of the inference-execution loop in different optimization settings. The black dashed box indicates the per-chunk latency.}
    \label{fig:dataflow}
    \vspace{-0.6em}
\end{figure}%
}

\vspace{-4pt}
\subsection{Basic Timing Model}
\vspace{-4pt}
\label{sec:timing-model}

In Figure~\ref{fig:dataflow}, let $T_{\mathrm{inf}}$ be the latency of full-model policy inference, $n$ the action chunk size~\cite{zhao2023learning},
and $T_{\mathrm{act}}$ the duration of one action.
In the synchronous full-model baseline, one chunk consists of one inference followed by $n$ executed actions.
We use this baseline chunk time as the reference cost:
\begin{equation}
    T_{\mathrm{chunk}}
    =
    T_{\mathrm{inf}} + nT_{\mathrm{act}} .
    \label{eq:chunk-time}
\end{equation}
\par\vspace{-6pt}

\paragraph{Policy-Intrinsic Optimization.}
Policy-intrinsic methods reduce computation inside the policy network.
We represent their timing effect by a latency ratio $\alpha \in (0,1]$.
This ratio can be changed by adjusting the number of quantization layers, the pruning ratio, and so on.
Under this class of methods, the modeled per-chunk time and per-chunk speedup are:
\begin{equation}
\begin{aligned}
    T_{\mathrm{chunk}}^{\mathrm{pl}}(\alpha)
    &= \alpha T_{\mathrm{inf}} + nT_{\mathrm{act}}, \quad
    \eta_{\mathrm{chunk}}^{\mathrm{pl}}(\alpha)
    = \frac{T_{\mathrm{chunk}}}{T_{\mathrm{chunk}}^{\mathrm{pl}}(\alpha)} .
\end{aligned}
    \label{eq:policy-cycle}
\end{equation}

The cost is that stronger lightweighting may reduce action quality.

\paragraph{Execution-Aware Optimization.}
Execution-aware methods like asynchronous inference change how inference is scheduled with robot execution,
and they usually require complex system scheduling.
To simplify the analysis, we performed abstract modeling of asynchronous inference by the emulating work~\cite{tang2025vlash}.
We summarize the scheduling knob by an action overlap depth $n' \in \{0,\ldots,n\}$, the number of primitive actions whose execution can overlap with the next inference.
The modeled per-chunk time and per-chunk speedup are:
\begin{equation}
\scalebox{0.92}{$\displaystyle
    T_{\mathrm{chunk}}^{\mathrm{ea}}(n')
    =
    T_{\mathrm{inf}} + nT_{\mathrm{act}}
    - \min\{T_{\mathrm{inf}}, n'T_{\mathrm{act}}\}, \quad
    \eta_{\mathrm{chunk}}^{\mathrm{ea}}(n')
    =
    \frac{T_{\mathrm{chunk}}}{T_{\mathrm{chunk}}^{\mathrm{ea}}(n')}.
$}
    \label{eq:ea-cycle}
\end{equation}
By overlapping the next inference with the current chunk's execution, each new action is computed from an observation taken before those actions complete,
which can introduce observation-execution misalignment and degrade action quality.

\subsection{Task-level Model}

\paragraph{End-to-End Time.}
A complete embodied task spans many action chunks.
The total number of chunks is not a fixed constant: it is an outcome of the rollout under the selected optimization.
For the baseline and the two optimization classes, the task completion times are:
\begin{equation}
\scalebox{0.92}{$\displaystyle
    T_{\mathrm{task}}^{\mathrm{base}} = N^{\mathrm{base}}T_{\mathrm{chunk}}, \quad
    T_{\mathrm{task}}^{\mathrm{pl}}(\alpha) = N^{\mathrm{pl}}(\alpha)T_{\mathrm{chunk}}^{\mathrm{pl}}(\alpha), \quad
    T_{\mathrm{task}}^{\mathrm{ea}}(n') = N^{\mathrm{ea}}(n')T_{\mathrm{chunk}}^{\mathrm{ea}}(n').
$}
\label{eq:task-time-speedup}
\end{equation}
Here $N^{\mathrm{base}}$, $N^{\mathrm{pl}}(\alpha)$, and $N^{\mathrm{ea}}(n')$ are the total numbers of chunks used by the corresponding rollouts.
The end-to-end speedup should compare per-task completion time against the baseline:
\begin{equation}
\scalebox{0.92}{$\displaystyle
    \eta_{\mathrm{e2e}}^{\mathrm{pl}}(\alpha)
    = \frac{T_{\mathrm{task}}^{\mathrm{base}}}{T_{\mathrm{task}}^{\mathrm{pl}}(\alpha)}
    = \frac{N^{\mathrm{base}}}{N^{\mathrm{pl}}(\alpha)}\eta_{\mathrm{chunk}}^{\mathrm{pl}}(\alpha), \quad
    \eta_{\mathrm{e2e}}^{\mathrm{ea}}(n')
    = \frac{T_{\mathrm{task}}^{\mathrm{base}}}{T_{\mathrm{task}}^{\mathrm{ea}}(n')}
    = \frac{N^{\mathrm{base}}}{N^{\mathrm{ea}}(n')}\eta_{\mathrm{chunk}}^{\mathrm{ea}}(n').
$}
\label{eq:e2e-speedup}
\end{equation}
$N(\cdot)$ is task- and policy-dependent and is measured from closed-loop rollouts.
Stronger lightweighting can increase $N$ through degraded action quality (policy-intrinsic optimization) or stale observations (execution-aware optimization).
Per-task time therefore need not track per-chunk time, and may even move in the opposite direction.

\vspace{-2pt}
\begin{tcolorbox}[colback=gray!8,colframe=black!70,boxrule=0.85pt,
    arc=2pt,left=6pt,right=6pt,top=3pt,bottom=3pt,
    fonttitle=\normalfont,coltitle=black]
\textbf{Overlooked Factor 1}: On static tasks, per-chunk speedup alone does not determine per-task speedup--degraded action quality can inflate the chunk count.
\end{tcolorbox}
\vspace{-2pt}

\paragraph{Action Delay.}
We also track the residual action delay, i.e., the time from an observation to the first action that can be executed from the corresponding policy output:
\begin{equation}
\begin{aligned}
    D_{\mathrm{act}}^{\mathrm{base}} &= T_{\mathrm{inf}}, \quad
    D_{\mathrm{act}}^{\mathrm{pl}}(\alpha) = \alpha T_{\mathrm{inf}}, \quad
    D_{\mathrm{act}}^{\mathrm{ea}}(n') = nT_{\mathrm{act}} + \max\{0, T_{\mathrm{inf}} - n'T_{\mathrm{act}}\}.
\end{aligned}
\label{eq:action-delay}
\end{equation}
Dynamic-task analysis further depends on $D_{\mathrm{act}}$, which captures the observation staleness accumulated by the time the corresponding action executes.
To cover both policy-intrinsic and execution-aware optimization, let $u$ denote the operating point of the chosen method: $u=\alpha$ for policy-intrinsic optimization and $u=n'$ for execution-aware optimization.
We use a structured surrogate to capture the two dominant factors in dynamic success:
  \begin{equation}
  \begin{aligned}
      \widehat{P}_{\mathrm{succ}}(u)
      &= \Phi\!\left(q(u), z(u)\right), \qquad
      z(u) = D_{\mathrm{act}}(u)\omega_{\mathrm{env}}, \qquad
      \partial \Phi / \partial q \ge 0, \qquad
      \partial \Phi / \partial z \le 0 .
  \end{aligned}
  \label{eq:obj-dynamic}
  \end{equation}
where $q(u)$ represents the action quality, $\omega_{\mathrm{env}}$ is the environment change rate (e.g., conveyor speed or moving-target velocity),
$z(u) = D_{\mathrm{act}}(u)\omega_{\mathrm{env}}$ measures how far the environment drifts during the action delay, and $\Phi$ maps these two factors to a success rate.
In this surrogate, we assume that dynamic success goes up when actions are more accurate, and goes down when the environment has drifted far from what the policy actually saw. As lightweighting strengthens,
$q(u)$ may fall (less computation or staler input) while $z(u)$ may also fall (faster inference shortens the observation-to-action gap).
The outcome for dynamic-task success therefore depends on the relative strength of these two trends.

\vspace{-2pt}
\begin{tcolorbox}[colback=gray!8,colframe=black!70,boxrule=0.85pt,
    arc=2pt,left=6pt,right=6pt,top=3pt,bottom=3pt,
    fonttitle=\normalfont,coltitle=black]
\textbf{Overlooked Factor 2}: On dynamic tasks, action quality alone does not determine task success--reduced observation staleness can outweigh the quality loss.
\end{tcolorbox}
\vspace{-2pt}

\paragraph{Compute Hardware.}
The timing terms above are not intrinsic to the policy alone.
$T_{\mathrm{inf}}$ is determined by the compute platform that runs the policy, whereas $T_{\mathrm{act}}$ is fixed by the robot execution side and the task setting. (1) For static tasks,
let $\rho=T_{\mathrm{inf}}/T_{\mathrm{act}}$ denote the inference-to-execution time ratio.
For a fixed task and a fixed optimization setting, the per-chunk speedups can be written as:
\begin{equation}
\begin{aligned}
    \eta_{\mathrm{chunk}}^{\mathrm{pl}}(\rho)
    &= \frac{\rho+n}{\alpha\rho+n}, \quad
    \eta_{\mathrm{chunk}}^{\mathrm{ea}}(\rho)
    = \frac{\rho+n}{\rho+n-\min\{\rho,n'\}} .
\end{aligned}
\label{eq:hardware-normalized-speedup}
\end{equation}
It means that the same lightweighting setting can yield different task-level effects on different compute platforms. (2) For dynamic tasks, the delay $D_{\rm act}$ can also be impacted by hardware capability, thereby affecting $z(u)$ and $\widehat{P}_{\mathrm{succ}}(u)$.

\vspace{-2pt}
\begin{tcolorbox}[colback=gray!8,colframe=black!70,boxrule=0.85pt,
    arc=2pt,left=6pt,right=6pt,top=3pt,bottom=3pt,
    fonttitle=\normalfont,coltitle=black]
\textbf{Overlooked Factor 3}: Across compute platforms, per-task trends can differ--the sweet spot may shift with the hardware's compute budget.
\end{tcolorbox}
\vspace{-2pt}

\endgroup

\vspace{-3pt}
\section{Experiments}
\label{sec:obs}
\vspace{-3pt}

Section~\ref{sec:unified-view} shows that inference latency is an incomplete proxy for embodied inference optimization.
In this section, we empirically characterize how this proxy breaks down through controlled sweeps of lightweighting across task regimes, policy models, optimization methods, and compute platforms.
These sweeps are designed to answer three questions: whether per-chunk speedup translates into end-to-end speedup in static tasks, whether reduced action delay translates into better accuracy in dynamic tasks, and whether the shape of the task-level trend and the location of its sweet spots are hardware-dependent.
The full experimental combinations, sweep ranges, and implementation details are reported in Appendix~\ref{app:simulation} and~\ref{app:realworld}.

\providecommand{\best}[1]{\textbf{#1}}
\providecommand{\accbest}[1]{\textbf{\textcolor{teal!70!black}{#1}}}
\providecommand{\fast}[1]{\underline{#1}}
\providecommand{\bad}[1]{\textcolor{red!70!black}{#1}}
\providecommand{\tablehead}{\rowcolor{black!12}}

\afterpage{%
\begin{figure*}[t]
\centering
\scriptsize
\setlength{\tabcolsep}{3.0pt}
\renewcommand{\arraystretch}{1.0}

\begin{minipage}[b]{0.69\textwidth}
\centering
\captionsetup[sub]{font=tiny, skip=1pt}

\subcaptionbox{\texttt{Cosmos-Policy @ RoboCasa @ PnP Sink To Counter @ quantization @ 6000 Ada}\label{tab:cosmos-pnp-transposed}}[0.49\linewidth]{%
\resizebox{\linewidth}{!}{%
\begin{tabular}{cccccc}
\toprule
\tablehead
Config & SR (\%) & $\alpha$ & $N$ & $T_{\rm chunk}$(s) & $T_{\rm task}$(s) \\
\midrule
fp16 & 64.0 & 1.00 & 15.94 & 0.993 & 15.83 \\
\rowcolor{black!4} w8a8 & 76.0 & 0.83 & 16.13 & 0.961 & 15.50 \\
w4-t1 & 66.0 & 0.82 & 16.09 & 0.958 & 15.42 \\
\rowcolor{black!4} w4-t2 & 40.0 & 0.79 & 15.85 & 0.953 & 15.10 \\
w4-t3 & 58.0 & 0.78 & 15.72 & 0.951 & \textbf{14.95} \\
\rowcolor{black!4} w4-t4 & 10.0 & 0.76 & 20.20 & 0.946 & 19.11 \\
w4-t5 & 0.0 & 0.73 & N.A. & 0.940 & N.A. \\
\bottomrule
\end{tabular}}}
\hfill
\subcaptionbox{\texttt{Cosmos-Policy @ RoboCasa @ Turn Off Microwave @ quantization @ 6000 Ada}\label{tab:cosmos-microwave-transposed}}[0.49\linewidth]{%
\resizebox{\linewidth}{!}{%
\begin{tabular}{cccccc}
\toprule
\tablehead
Config & SR (\%) & $\alpha$ & $N$ & $T_{\rm chunk}$(s) & $T_{\rm task}$(s) \\
\midrule
fp16 & 100.0 & 1.00 & 14.10 & 0.993 & \textbf{14.00} \\
\rowcolor{black!4} w8a8 & 100.0 & 0.83 & 16.24 & 0.961 & 15.61 \\
w4-t1 & 100.0 & 0.82 & 15.88 & 0.958 & 15.21 \\
\rowcolor{black!4} w4-t2 & 68.0 & 0.79 & 18.06 & 0.953 & 17.20 \\
w4-t3 & 70.0 & 0.78 & 19.06 & 0.951 & 18.12 \\
\rowcolor{black!4} w4-t4 & 12.0 & 0.76 & 24.00 & 0.946 & 22.71 \\
w4-t5 & 0.0 & 0.73 & N.A. & 0.940 & N.A. \\
\bottomrule
\end{tabular}}}

\vspace{0.15em}

\subcaptionbox{\texttt{LingBot-VA @ RoboTwin 2.0 @ stack bowls\_3 @ pruning @ 6000 Ada}\label{tab:lingbot-va-stack-transposed}}[0.49\linewidth]{%
\resizebox{\linewidth}{!}{%
\begin{tabular}{cccccc}
\toprule
\tablehead
Config & SR (\%) & $\alpha$ & $N$ & $T_{\rm chunk}$(s) & $T_{\rm task}$(s) \\
\midrule
base & 90.0 & 1.00 & 14.87 & 2.165 & 32.19 \\
\rowcolor{black!4} p1 & 86.0 & 0.98 & 14.88 & 2.129 & 31.68 \\
p2 & 84.0 & 0.88 & 15.29 & 1.940 & 29.66 \\
\rowcolor{black!4} p3 & 76.0 & 0.88 & 15.05 & 1.942 & \textbf{29.23} \\
p4 & 34.0 & 0.89 & 22.18 & 1.958 & 43.42 \\
\bottomrule
\end{tabular}}}
\hfill
\subcaptionbox{\texttt{pi0.5 @ LIBERO @ 4 task avg @ async @ AGX 50W}\label{tab:vlash-original-transposed}}[0.49\linewidth]{%
\resizebox{\linewidth}{!}{%
\begin{tabular}{cccccc}
\toprule
\tablehead
Config & SR (\%) & $n'$ & $N$ & $T_{\rm chunk}$(s) & $T_{\rm task}$(s) \\
\midrule
d0 & 93.0 & 0 & 51.4 & 3.255 & 167.16 \\
\rowcolor{black!4} d1 & 94.4 & 1 & 49.8 & 3.222 & \textbf{160.61} \\
d2 & 93.6 & 2 & 51.6 & 3.189 & 164.43 \\
\rowcolor{black!4} d3 & 92.8 & 3 & 53.8 & 3.155 & 169.80 \\
d4 & 84.4 & 4 & 57.3 & 3.122 & 178.89 \\
\bottomrule
\end{tabular}}}
\end{minipage}
\hfill
\begin{minipage}[b]{0.30\textwidth}
\centering
\captionsetup[sub]{font=tiny, skip=1pt}

\subcaptionbox{Trajectory\label{fig:obs1-traj-transposed}}[\linewidth]{%
    \includegraphics[
        width=0.8\linewidth,
        height=1.26in,
        keepaspectratio
    ]{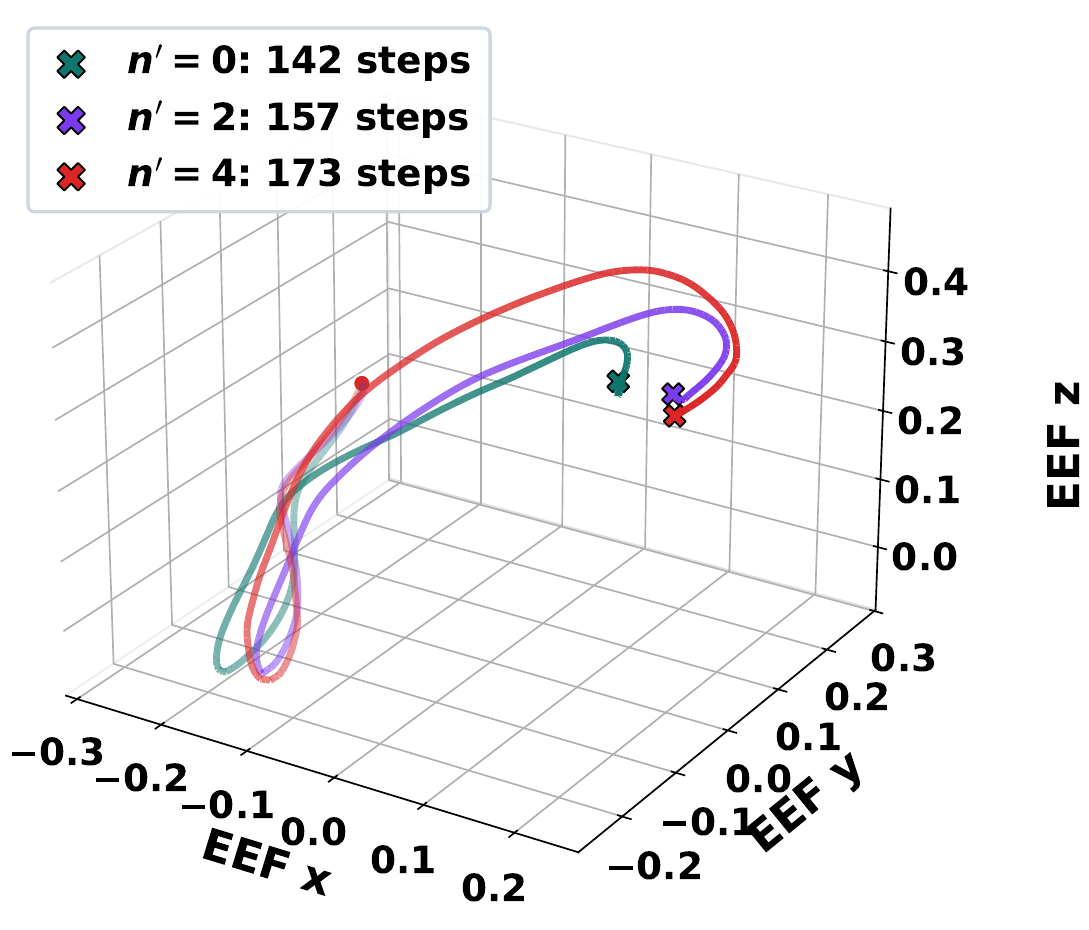}
}

\vspace{2pt}

\subcaptionbox{Key frames\label{fig:obs1-frames-transposed}}[\linewidth]{%
    \includegraphics[
        width=0.96\linewidth,
        height=1.26in,
        keepaspectratio
    ]{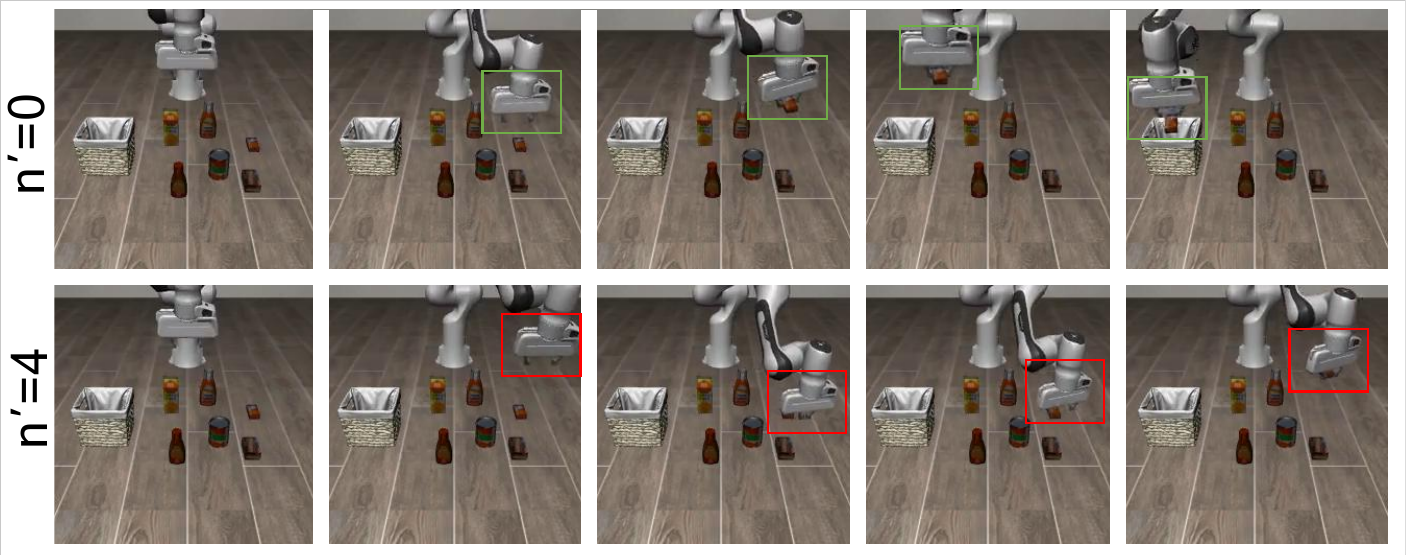}
}
\end{minipage}

\caption{Static-task results under \texttt{policy model @ task @ sub-task @ optimization method @ hardware} settings. (a)--(d) report SR, optimization strength, $N$, $T_{\rm chunk}$, and $T_{\rm task}$. To avoid interference from timeouts in failed tasks, the reported number $N$ is obtained by averaging over all \textbf{successful trials}. For specific configuration details, see Appendix~\ref{app:simulation-optimization}. (e) and (f) visualize the simulation case in (d).}
\label{fig:obs1}
\vspace{-0.5em}
\end{figure*}%
}

\vspace{-2pt}
\subsection{Setup}
\label{sec:obs-setup}
\vspace{-4pt}
\paragraph{Policy Models.}
We evaluate policy models at different scales.
We use MLP-Mixer~\cite{tolstikhin2021mlpmixerallmlparchitecturevision, tang2025vlash} and diffusion policy~\cite{chi2024diffusionpolicyvisuomotorpolicy} models for simple control tasks, and we also use foundation-scale embodied models in complex environment tasks,
including VLAs such as $\pi_{0.5}$~\cite{black2025pi0.5} and DynamicVLA~\cite{xie2026dynamicvla}, and WAMs such as Cosmos-Policy~\cite{kim2026cosmos} and LingBot-VA~\cite{li2026causal}.
\vspace{-8pt}
\paragraph{Optimization Methods.}
For policy-intrinsic optimization, we use quantization, pruning, and sampling-step reduction to accelerate the model inference.
To characterize task-level trends with fine granularity, we construct a series of progressively lightweight operating points for each method to adjust $\alpha$.
For execution-aware optimization, we follow the asynchronous inference settings in VLASH~\cite{tang2025vlash} and progressively vary the overlap depth $n'$.
\vspace{-8pt}
\paragraph{Compute Platforms.}
We evaluate the latency of policy models' inference on edge devices such as the Jetson AGX Orin~\cite{nvidiaNVIDIAJetson} (denoted \texttt{AGX 15W/30W/50W}) and server-class GPUs, including the RTX 3090~\cite{nvidia30903090}, RTX 4070~\cite{nvidia4070} and the 6000 Ada~\cite{nvidiaNVIDIA6000}. The power of edge devices can be adjusted to achieve different levels of computational capability.
\vspace{-8pt}
\paragraph{Simulation Task Setup.}
We evaluate both task regimes defined in Section~\ref{sec:unified-view}. For static tasks, we use LIBERO~\cite{liu2023libero}, RoboCasa~\cite{robocasa2024}, and RoboTwin-2.0~\cite{chen2025robotwin2}.
These benchmarks primarily involve manipulation tasks such as grasping and placing stationary objects, where the task-relevant object states remain effectively unchanged during policy inference.
For dynamic tasks, we use Kinetix~\cite{matthews2024kinetix} and DOM~\cite{xie2026dynamicvla}.
These benchmarks' task-relevant states continue to evolve during policy inference, and their latency-injection protocols allow us to emulate different inference delays.
\vspace{-8pt}
\paragraph{Real-World Task Setup.}
We further evaluate our observations on a real UR5e robotic arm~\cite{universalrobotsESeriesRobots}.
In the static setting, the robot needs to pick up a cube on the tabletop and place it in a target box.
In the dynamic setting, the robot needs to grasp a cube moving on a conveyor. These two settings share the same manipulation objective but differ in whether the task-relevant object state evolves during policy inference.

\vspace{-3pt}
\subsection{Empirical Finding 1: A Per-Chunk Speedup Need Not Speed Up the Static Task}
\vspace{-4pt}
\label{sec:obs1}

As shown in Figure~\ref{fig:obs1}, across all four model–task pairs, lightweighting--policy-intrinsic ($\alpha$) or execution-aware ($n'$)--generally reduces the per-chunk time $T_{\rm chunk}$, but action quality also declines, so the chunk count $N$ grows. Their product $T_{\rm task}=N\cdot T_{\rm chunk}$ thus depends on which trend dominates.
In Figure~\ref{fig:obs1}(a), (c), and (d), $N$ stays nearly flat under mild lightweighting and inflates sharply past a task-specific operating point, so $T_{\rm task}$ traces a U-shape with an interior minimum below the baseline.
Figure~\ref{fig:obs1}(b) is the corner case: $N$ inflates from the very first optimization step, so no setting beats the baseline.
Figure~\ref{fig:obs1}(e) and~(f) visualize the underlying mechanism on a representative asynchronous trial: as $n'$ grows, the robot arm traces longer end-effector paths and makes more failed grasp attempts before completing the task.

\begin{tcolorbox}[colback=blue!5,colframe=blue!45!black,boxrule=0.85pt,
    arc=2pt,left=6pt,right=6pt,top=3pt,bottom=3pt,
    fonttitle=\normalfont,coltitle=black]
\textbf{Empirical Finding 1}: On static tasks, $T_{\rm task}$ is non-monotonic in lightweighting strength--an interior sweet spot may exist between two regimes: mild lightweighting reduces $T_{\rm task}$ via per-chunk speedup, aggressive lightweighting inflates the chunk count $N$ and reverses the gain.
\end{tcolorbox}

\vspace{-3pt}
\subsection{Empirical Finding 2: An Imprecise but Faster Reaction Can Improve the Dynamic Task}
\vspace{-4pt}
\label{sec:obs2}

\begin{figure*}[t]
    \centering

    \captionsetup{skip=4pt}
    \captionsetup[sub]{font=scriptsize,skip=2pt}

    \definecolor{localBlue}{HTML}{1F77B4}
    \definecolor{agxGreen}{HTML}{2CA02C}

    \newcommand{\tableAwidthScale}{0.5}
    \newcommand{\tableAheightScale}{0.55}
    \newcommand{\tableBwidthScale}{0.5}
    \newcommand{\tableBheightScale}{0.55}

    \newcommand{\imageCwidthScale}{1.0}
    \newcommand{\imageCheightScale}{3.3cm}

    \newcommand{\tableFontSize}{\normalsize}

    \begin{minipage}[t]{0.99\textwidth}
        \centering

        \begin{subfigure}[t]{0.42\linewidth}
            \centering
            {\tiny
            \setlength{\tabcolsep}{1.5pt}
            \renewcommand{\arraystretch}{1.05}
                \begin{tabular}{lrrrrrr}
                    \toprule
                    \tablehead
                    setting &
                    car &
                    catapult &
                    catcher &
                    grasp &
                    lunar &
                    cheetah \\
                    \midrule
                    \multicolumn{7}{l}{\textit{Ideal -- zero inference delay}} \\
                    \texttt{inf-hw s1}          & 0.984 & 0.555 & 0.953 & 0.984 & 0.914 & 0.672 \\
                    \texttt{inf-hw s3}          & \textbf{1.000} & \textbf{0.727} & \textbf{0.984} & \textbf{0.984} & 0.914 & 0.836 \\
                    \texttt{inf-hw s5}          & 0.992 & 0.680 & 0.969 & 0.953 & \textbf{0.922} & \textbf{0.898} \\
                    \cmidrule(lr){1-7}
                    \multicolumn{7}{l}{\textit{AGX 15W Simulation-- with measured delay}} \\
                    \texttt{agx15 s1 d0}        & \textbf{0.984} & \textbf{0.555} & \textbf{0.953} & \textbf{0.984} & 0.914 & 0.672 \\
                    \texttt{agx15 s3 d1}        & 0.938 & 0.461 & 0.586 & 0.844 & \textbf{0.914} & 0.836 \\
                    \texttt{agx15 s5 d2}        & 0.820 & 0.477 & 0.500 & 0.805 & 0.867 & \textbf{0.844} \\
                    \bottomrule
                \end{tabular}
            }
            \caption{\texttt{MLP-Mixer @ Kinetix}}
        \end{subfigure}\hspace{0.4em}%
        \begin{subfigure}[t]{0.25\linewidth}
            \centering
            {\tiny
            \setlength{\tabcolsep}{1.2pt}
            \renewcommand{\arraystretch}{1.05}
                \begin{tabular}{crrr}
                    \toprule
                    \tablehead
                    speed & steps & $D_{\rm act}$(ms) & SR(\%) \\
                    \midrule
                    \multirow{4}{*}{100\%} & 1  & 14.28 & 68 \\
                                           & 2  & 16.25 & \textbf{76} \\
                                           & 5  & 19.92 & 74 \\
                                           & 10 & 25.69 & 70 \\
                    \midrule
                    \multirow{4}{*}{125\%} & 1  & 14.28 & 36 \\
                                           & 2  & 16.25 & 34 \\
                                           & 5  & 19.92 & \textbf{48} \\
                                           & 10 & 25.69 & 46 \\
                    \bottomrule
                \end{tabular}
            }
            \caption{\texttt{DynamicVLA @ DOM-CR}}
        \end{subfigure}\hfill
        \begin{subfigure}[t]{0.31\linewidth}
            \centering
            \begin{minipage}[c][\imageCheightScale][c]{\linewidth}
                \centering
                \includegraphics[width=\linewidth]{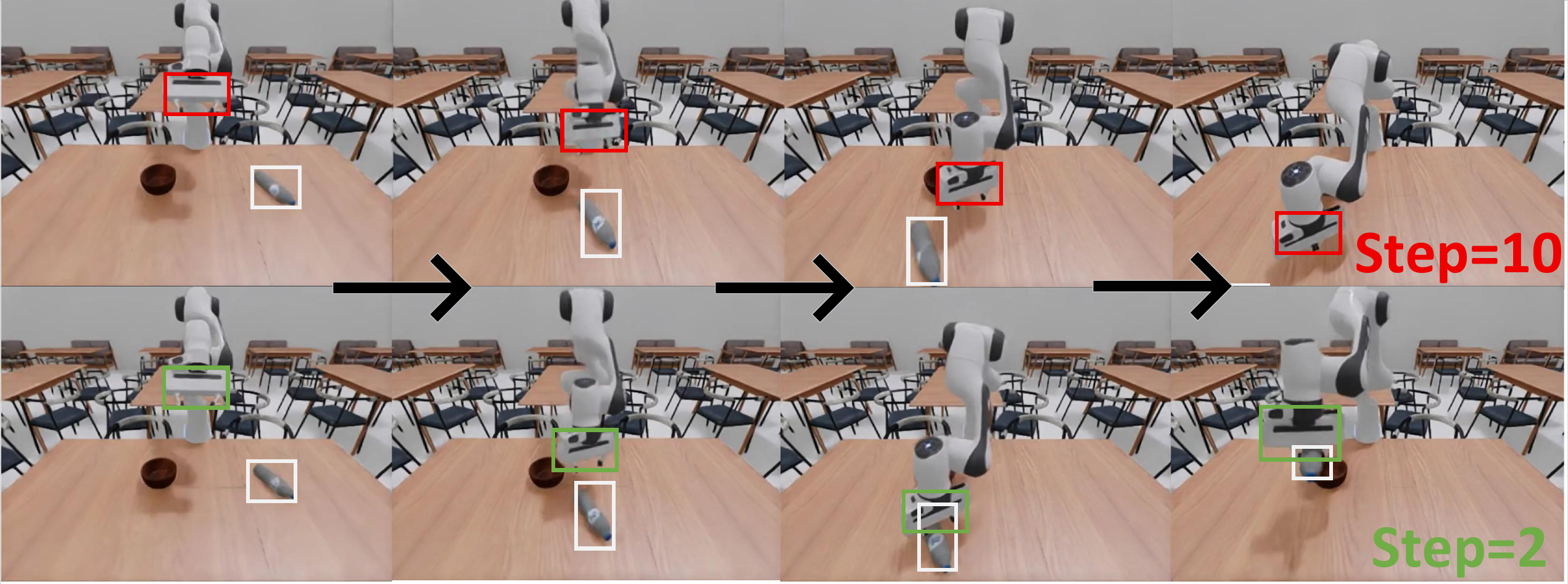}
            \end{minipage}
            \caption{DOM-CR Visualization}
        \end{subfigure}

    \end{minipage}
    \caption{(a) Accuracy on Kinetix's sub-tasks. \texttt{inf-hw} assumes that policy inference is instantaneous (i.e., zero delay). \texttt{s[step]} denotes the number of sampling steps.
In actual deployment, we measured the policy's inference speed on AGX 15W device and converted its inference delay into an equivalent number of delay steps \texttt{d[delay]} in the simulator as shown in Appendix~\ref{app:simulation-tasks}.
(b) Accuracy on the DOM-CR sub-task. ``speed" indicates the motion speed of objects in the task relative to the original object speed in the dataset.
(c) DOM-CR simulator visualization.}
    \label{fig:obs2}
    \vspace{-0.3em}
\end{figure*}

On dynamic tasks, lightweighting--policy-intrinsic ($\alpha$) or execution-aware ($n'$)--drives two competing trends: action quality $q$ falls because computation drops, but observation staleness $z = D_{\rm act}\omega_{\rm env}$ also falls because the policy reacts on fresher observations. Task success rate $P_{\rm succ}$ thus depends on which trend dominates.
Figure~\ref{fig:obs2}(a) (Kinetix) shows the regime flip: under the zero-delay assumption (\texttt{inf-hw}), accuracy peaks usually at the larger sampling-step $s$; once realistic delay is injected to simulate AGX~15W's inference latency, smaller $s$ becomes the better choice because the reaction-speed gain outweighs the action-quality loss.
Figure~\ref{fig:obs2}(b) (DOM-CR) sweeps both $s$ and the environment speed $\omega_{\rm env}$: at every speed, the best configuration is neither the lowest-delay nor the finest-action one, but an interior sweet spot that balances $q$ and $z$; the sweet-spot location itself shifts with $\omega_{\rm env}$.
Figure~\ref{fig:obs2}(c) visualizes the paradox in a DOM-CR scene.

\vspace{-3pt}
\begin{tcolorbox}[colback=blue!5,colframe=blue!45!black,boxrule=0.85pt,
    arc=2pt,left=6pt,right=6pt,top=3pt,bottom=3pt,
    fonttitle=\normalfont,coltitle=black]
\textbf{Empirical Finding 2}: On dynamic tasks, task success rate $P_{\rm succ}$ is non-monotonic in lightweighting strength--an interior sweet spot may exist between two regimes: mild lightweighting raises $P_{\rm succ}$ by reducing observation staleness $z$, aggressive lightweighting collapses action quality $q$ and reverses the gain.
\end{tcolorbox}
\vspace{-3pt}

\vspace{-3pt}
\subsection{Empirical Finding 3: The Trend and Optimum Shifts with Hardware}
\vspace{-3pt}
\label{sec:obs3}

\begin{figure}[t]
    \centering
    \captionsetup{skip=4pt}
    \begin{subfigure}[t]{0.33\linewidth}
        \centering
        \includegraphics[width=\linewidth]{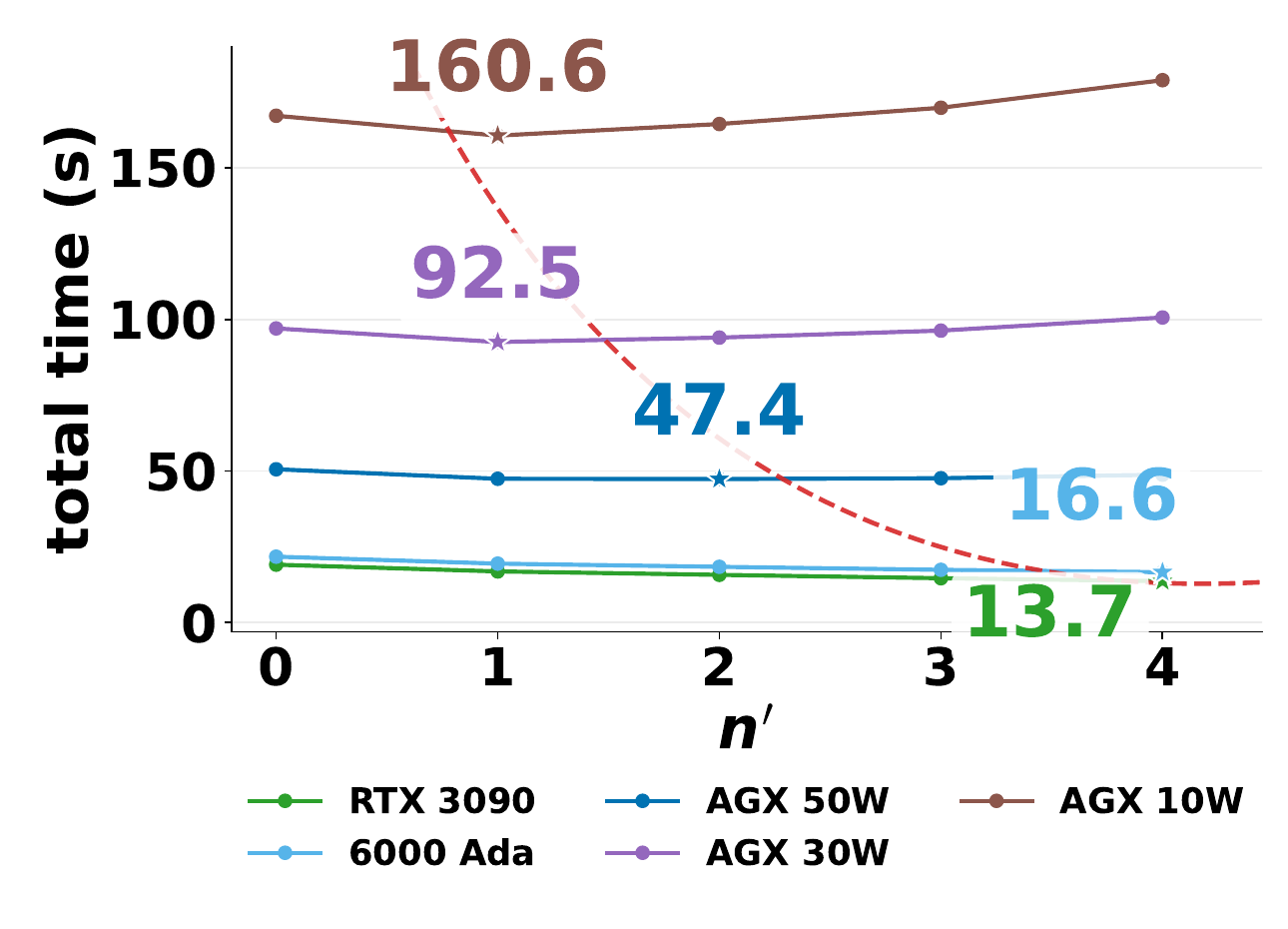}
        \caption{Static task}
    \end{subfigure}\hfill
    \begin{subfigure}[t]{0.65\linewidth}
        \centering
        \includegraphics[width=\linewidth]{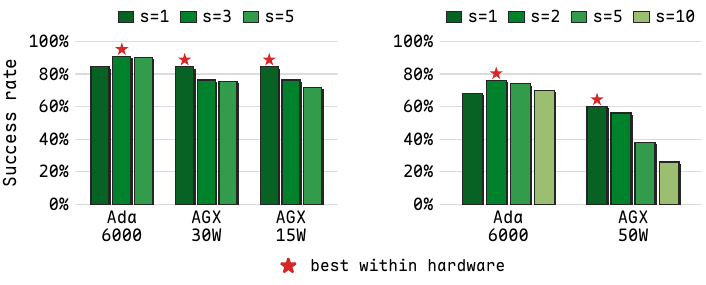}
        \caption{Dynamic task}
    \end{subfigure}
    \caption{(a) On different hardware platforms, the average time required for pi0.5 to complete the LIBERO task versus the asynchronous overlap depth $n^\prime$ , with the red dashed line indicating how its sweet spot shifts with hardware; (b) On different hardware platforms, the task accuracy of Kinetix (left) and DOM CR (right) versus the sampling steps, where the location of highest accuracy has shifted.}
    \label{fig:obs3}
    \vspace{-0.3em}
\end{figure}

Empirical Findings~1 and~2 established that a sweet spot may exist on both static and dynamic tasks. We next ask whether its location is hardware-dependent.
To isolate the hardware effect, we fix the task and the lightweighting sweep, and vary only the inference hardware, which changes the inference-to-execution ratio $\rho = T_{\rm inf}/T_{\rm act}$.
Figure~\ref{fig:obs3}(a) (static, async sweep over $n'$): on high-performance hardware ($\rho$ small), the per-chunk speedup $\eta_{\rm chunk}$ is large enough that the gain from inference optimization outweighs the chunk-count inflation, so the sweet spot drifts toward more aggressive overlap; on low-performance hardware ($\rho$ large), the per-chunk gain shrinks while $N$ inflates rapidly, so the sweet spot retreats to milder overlap.
Figure~\ref{fig:obs3}(b) (dynamic, sampling-step sweep over $s$): on high-performance hardware, inference delay $D_{\rm act}$ is no longer the dominant factor limiting accuracy, so the sweet spot favors larger $s$ where the policy retains higher action quality $q$; on low-performance hardware, observation staleness $z$ dominates and the policy must trade off action quality against reaction speed, so the sweet spot shifts to smaller $s$.
Across both regimes, the sweet-spot location is qualitatively sensitive to the hardware compute budget; the quantitative derivation is in Appendix~\ref{app:sweetspot-rho-drift}.

\begin{tcolorbox}[colback=blue!5,colframe=blue!45!black,boxrule=0.85pt,
    arc=2pt,left=6pt,right=6pt,top=4pt,bottom=4pt,
    fonttitle=\normalfont,coltitle=black]
\textbf{Empirical Finding 3}: Across compute platforms, the per-task sweet spot shifts with the hardware's compute budget that affects $\rho$.
\end{tcolorbox}

\raggedbottom
\afterpage{\flushbottom}
\vspace{-2pt}
\subsection{Case Study: Predictive TISED on Dynamic Task}
\label{sec:predictive-case}
\vspace{-3pt}

Building on the trade-off between action quality and reaction speed,
we use TISED to compose separately calibrated responses and predict task-level performance across hardware platforms.
As shown in Figure~\ref{fig:predictive-case}, taking Kinetix~\cite{matthews2024kinetix} as an example, we measure and fit the \textit{hardware latency response} $f_{\mathrm{HW}}$, \textit{zero-delay policy response} $g$, and \textit{fixed-policy delay response} $h$.
We approximate $\Phi$ in Equation~\ref{eq:obj-dynamic} by multiplying the quality and delay responses to obtain the \textit{composed task response}, and compare predicted and measured success rates across hardware platforms.
The composed predictions capture the observed performance trends and hardware-dependent shift of the sweet spot, highlighting the importance of jointly accounting for policy quality and inference latency.
Experiment settings and fitting details are provided in Appendix~\ref{app:simulation-tasks}.

\begingroup
\setlength{\parskip}{0pt}
\par\vspace{4pt}\noindent
\begin{minipage}{\linewidth}
    \centering
    \includegraphics[width=\linewidth]{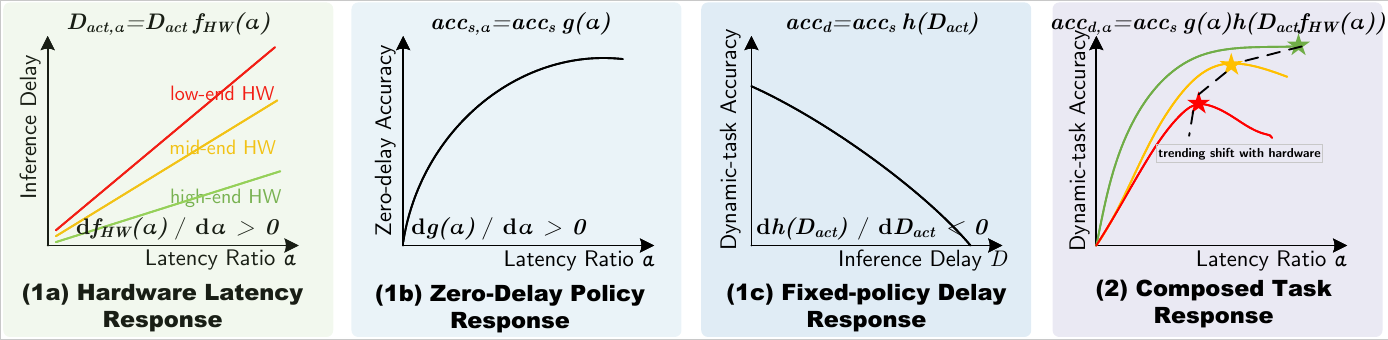}\\[4pt]
    \includegraphics[width=\linewidth]{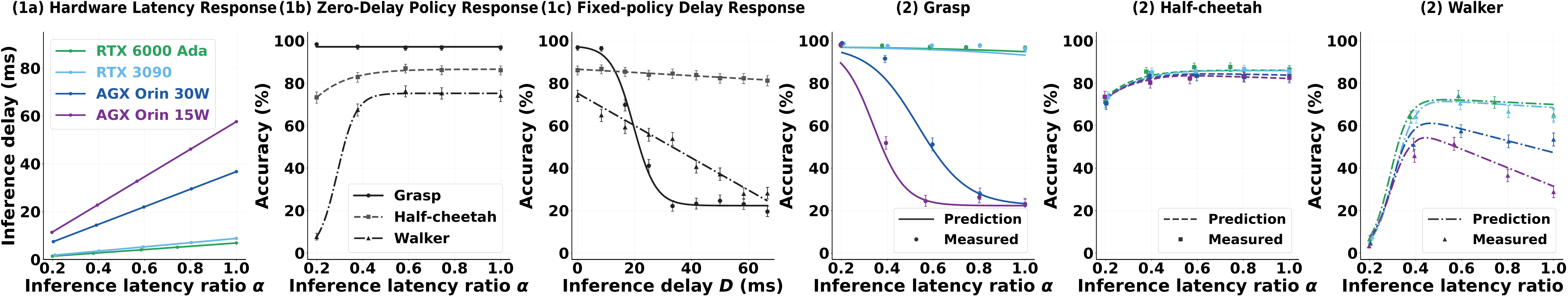}
    \captionsetup{type=figure,skip=2pt}
    \caption{Predictive TISED. Top: schematic illustration of response decomposition and composition. Bottom: experimental results showing calibrated responses and predicted versus measured success rates on three Kinetix tasks: \textit{Grasp} is mainly sensitive to delay, \textit{Half-cheetah} to lightweighting, and \textit{Walker} to both. Curves and markers in the right three panels denote predictions and measurements, respectively.}
    \label{fig:predictive-case}
\end{minipage}
\par
\endgroup

\subsection{Real-World Experiments}
\vspace{-0.6em}
\label{sec:real-world}
We conduct two real-world studies on a UR5e robot arm as shown in Figure~\ref{fig:real-world-results}.
For the static task,
we use a fine-tuned $\pi_{0.5}$ model~\cite{black2025pi0.5} and measure $T_{\rm inf}$ and $T_{\rm task}$ over all successful trials under different sampling-step settings.
Table~\ref{tab:real-static} shows that the two metrics exhibit opposite trends as lightweighting becomes more aggressive.

For the dynamic experiment, we use a fine-tuned diffusion policy model~\cite{chi2024diffusionpolicyvisuomotorpolicy} and compute the score based on gripper--cube contact.
Table~\ref{tab:real-dynamic} shows that diffusion policies with fewer sampling steps can achieve higher dynamic-task scores, because their faster reaction outweighs the loss in action quality.
Tables~\ref{tab:step1_5_7_10} and~\ref{tab:step4_16} further show that the sweet-spot step in both regimes shifts across hardware.
Figures~\ref{fig:grasp-distance-a} and~\ref{fig:grasp-distance-b} visualize this static-task paradox at the trajectory level: as the sampling step decreases, the gripper width trace shows visibly more oscillation.
Together, these two real-world studies provide physical evidence for the speedup paradox and corresponding hardware shift.

\par\begingroup\setlength{\parskip}{0pt}
\newlength{\realworldsceneheight}
\setlength{\realworldsceneheight}{0.144\linewidth}
\vspace{4pt}\nointerlineskip\noindent
    \begin{minipage}{\linewidth}
    \captionsetup{type=figure}
    \captionsetup{skip=4pt}
    \captionsetup[sub]{font=scriptsize,skip=2pt}
    \centering
    \begin{minipage}[c]{0.30\linewidth}
        \centering
        \begin{subfigure}[b]{0.48\linewidth}
            \centering
            {\setlength{\fboxsep}{0.6pt}%
            \begin{overpic}[width=\linewidth]{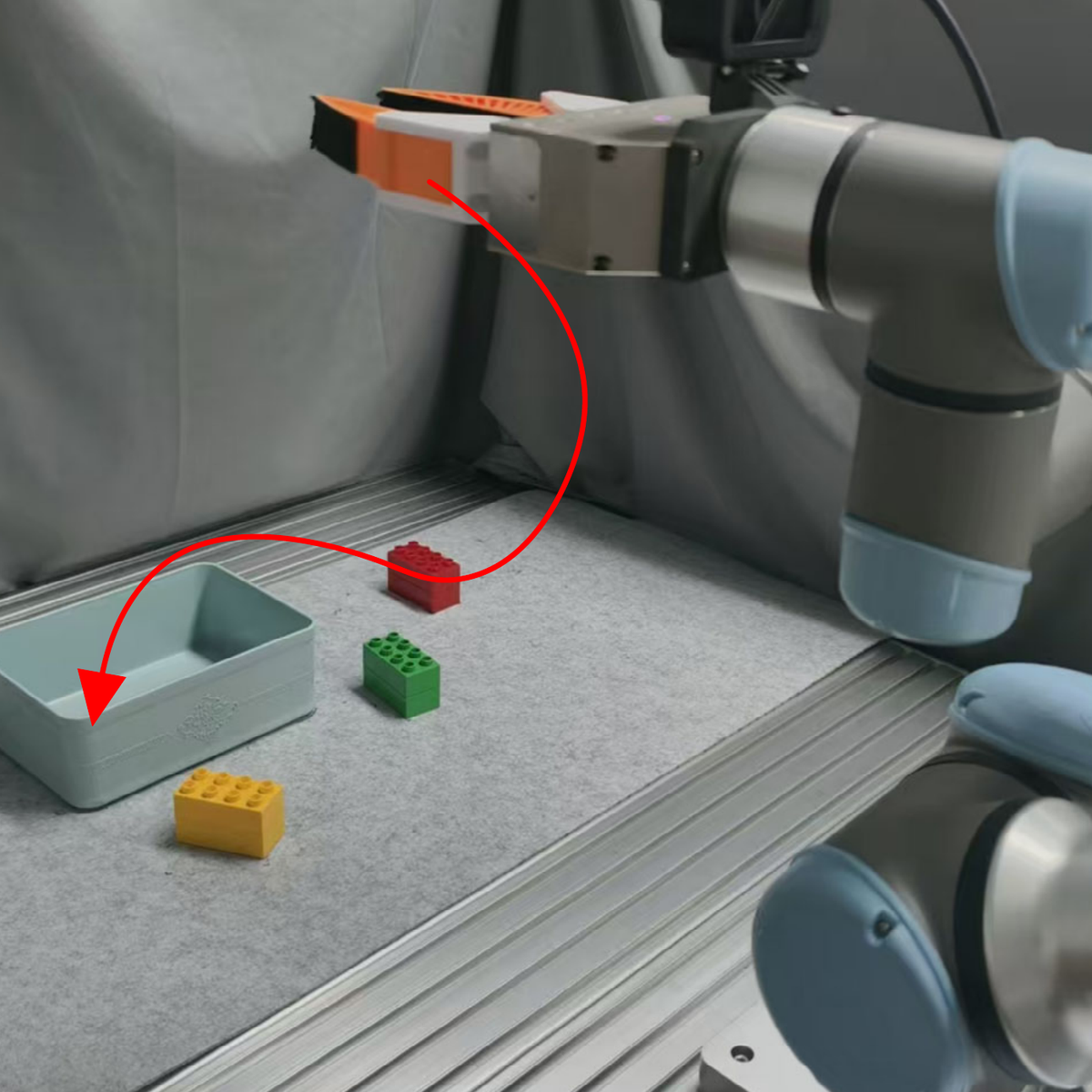}
                \put(3,59){\colorbox{white}{\color{red}\sffamily\bfseries\fontsize{5}{5.5}\selectfont long distance}}
            \end{overpic}}
            \caption{Static setting}
            \label{fig:real-static-scene}
        \end{subfigure}\hfill
        \begin{subfigure}[b]{0.48\linewidth}
            \centering
            {\setlength{\fboxsep}{0.6pt}%
            \newcommand{\dynamicscorelabel}[2]{%
                {\setlength{\fboxsep}{0.35pt}%
                \colorbox{white}{\color[HTML]{#1}\sffamily\bfseries\fontsize{5}{5.5}\selectfont Score~#2}}}
            \begin{overpic}[width=\linewidth]{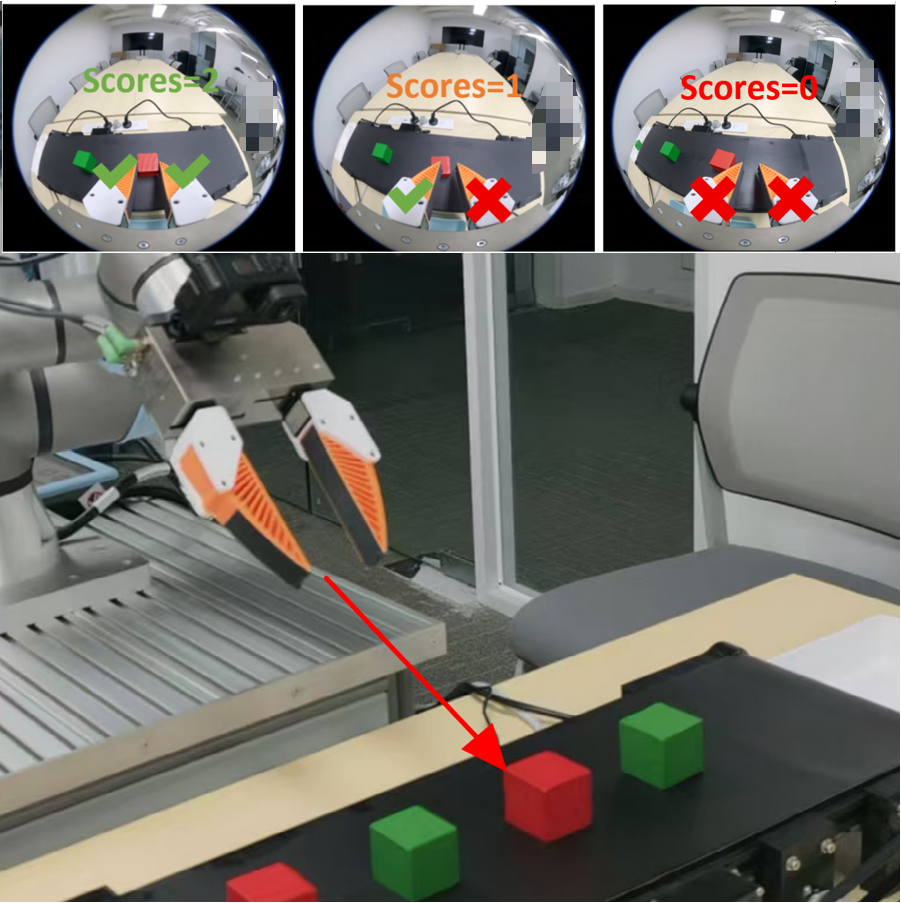}
                \put(16.7,91.1){\makebox(0,0){\dynamicscorelabel{70AD47}{2}}}
                \put(50.0,91.1){\makebox(0,0){\dynamicscorelabel{ED7D31}{1}}}
                \put(83.3,91.1){\makebox(0,0){\dynamicscorelabel{FF0000}{0}}}
                \put(37,33){\colorbox{white}{\color{red}\sffamily\bfseries\fontsize{5}{5.5}\selectfont short distance}}
            \end{overpic}}
            \caption{Dynamic setting}
            \label{fig:real-dynamic-scene}
        \end{subfigure}
    \end{minipage}\hfill
    \begin{minipage}[c]{0.68\linewidth}
        \centering
        \begin{subfigure}[b]{0.48\linewidth}
            \centering
            \includegraphics[width=\linewidth,height=\realworldsceneheight]{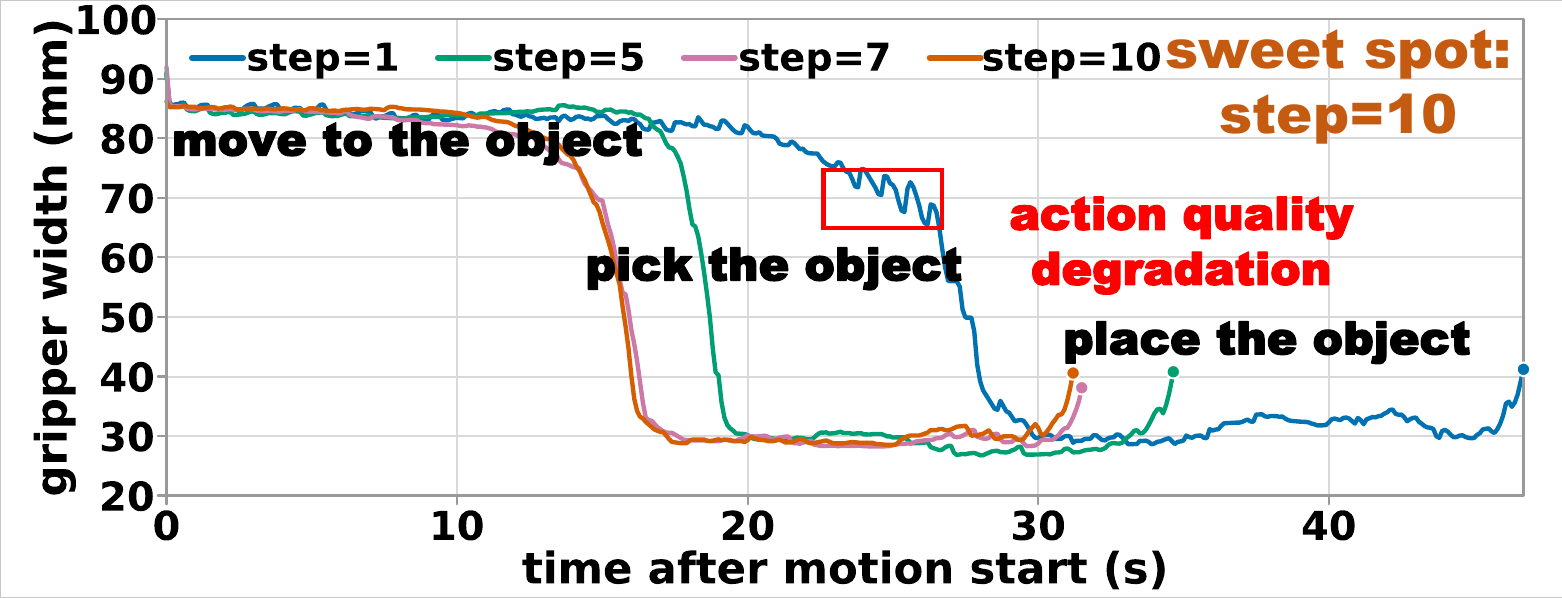}
            \caption{Gripper width visualization (fast HW)}
            \label{fig:grasp-distance-a}
        \end{subfigure}\hfill
        \begin{subfigure}[b]{0.48\linewidth}
            \centering
            \includegraphics[width=\linewidth,height=\realworldsceneheight]{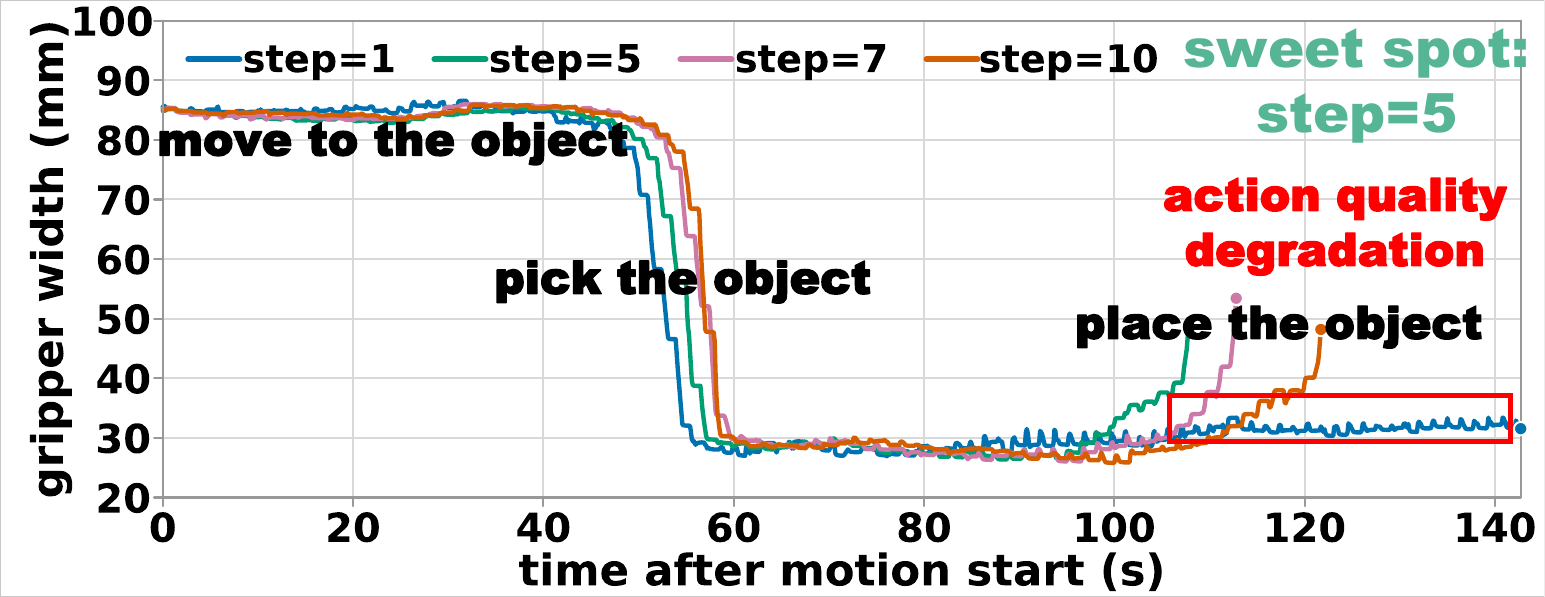}
            \caption{Gripper width visualization (slow HW)}
            \label{fig:grasp-distance-b}
        \end{subfigure}
    \end{minipage}
    \caption{Real-world experiment visualization on the UR5e platform. (a)(b) Task setups (static pick-and-place; dynamic conveyor grasp). (c)(d) Static-task gripper width under varying step counts with corresponding inference latencies on fast (6000 Ada) and slow (AGX 50W) hardware.}
    \label{fig:real-world-results}
    \end{minipage}
    \par
\endgroup

\par\begingroup\setlength{\parskip}{0pt}
\vspace{4pt}\nointerlineskip\noindent
    \begin{minipage}{\linewidth}
    \captionsetup{type=table}
    \captionsetup[sub]{font=scriptsize,skip=2pt,labelformat=parens}
    \centering
    \newlength{\figSixTabH}\setlength{\figSixTabH}{11mm}
    \subcaptionbox{Static paradox\label{tab:real-static}}[0.25\linewidth]{%
        \begin{minipage}[c][\figSixTabH][c]{\linewidth}\centering
        {\fontsize{6.5pt}{7.5pt}\selectfont
        \setlength{\tabcolsep}{0.6pt}
        \renewcommand{\arraystretch}{0.96}
        \begin{tabular}{ccccc}
            \hline
            \tablehead
            Config & $T_{\mathrm{inf}}$(s) & $N$ & $T_{\mathrm{task}}$(s) & Succ. \\
            \hline
            step=1  & \textbf{0.162} & 57.4 & 45.42 & 5/5 \\
            \rowcolor{black!4} step=7  & 0.186 & 49.2 & 38.74 & 5/5 \\
            step=10 & 0.199 & \textbf{47.2} & \textbf{37.26} & 5/5 \\
            \hline
        \end{tabular}
        }
        \end{minipage}%
    }\hfill
    \subcaptionbox{Static drift\label{tab:step1_5_7_10}}[0.24\linewidth]{%
        \begin{minipage}[c][\figSixTabH][c]{\linewidth}\centering
        {\fontsize{6.5pt}{7.5pt}\selectfont
        \setlength{\tabcolsep}{0.6pt}
        \renewcommand{\arraystretch}{0.96}
        \begin{tabular}{ccccc}
            \hline
            \tablehead
            HW & s=1 & s=5 & s=7 & s=10 \\
            \hline
            RTX 4070 & 44.6 & 41.5 & 38.6 & \textbf{37.1} \\
            \rowcolor{black!4} AGX 50W  & 159.5 & \textbf{108.5} & 116.9 & 119.2 \\
            \hline
        \end{tabular}
        }
        \end{minipage}%
    }\hfill
    \subcaptionbox{Dynamic paradox\label{tab:real-dynamic}}[0.25\linewidth]{%
        \begin{minipage}[c][\figSixTabH][c]{\linewidth}\centering
        {\fontsize{6.5pt}{7.5pt}\selectfont
        \setlength{\tabcolsep}{0.6pt}
        \renewcommand{\arraystretch}{0.96}
        \begin{tabular}{ccccc}
            \hline
            \tablehead
            Config & $D_{\rm act}$(s) & Pos. 1 & Pos. 2 & Pos. 3 \\
            \hline
            step=16 & 0.77 & 10/20 & 7/20 & 9/20 \\
            \rowcolor{black!4} step=4  & 0.41 & \textbf{16/20} & \textbf{13/20} & \textbf{16/20} \\
            \hline
        \end{tabular}
        }
        \end{minipage}%
    }\hfill
    \subcaptionbox{Dynamic drift\label{tab:step4_16}}[0.25\linewidth]{%
        \begin{minipage}[c][\figSixTabH][c]{\linewidth}\centering
        {\fontsize{6.5pt}{7.5pt}\selectfont
        \setlength{\tabcolsep}{0.6pt}
        \renewcommand{\arraystretch}{0.96}
        \begin{tabular}{ccccc}
            \hline
            \tablehead
            HW & s=1 & s=2 & s=4 & s=16 \\
            \hline
            RTX 4070 & 0/60 & 20/60 & 48/60 & \textbf{50/60} \\
            \rowcolor{black!4} AGX 30W & 0/60 & 15/60 & \textbf{45/60} & 26/60 \\
            \hline
        \end{tabular}
        }
        \end{minipage}%
    }
    \caption{Real-world results on the UR5e platform.
    (a) Static \emph{paradox} (Empirical Finding~1): lower inference latency yields slower per-task time (RTX 4070).
    (b) Static \emph{drift} (Empirical Finding~3): sweet step shifts from 10 (RTX 4070) to 5 (AGX 50W).
    (c) Dynamic \emph{paradox} (Empirical Finding~2): fewer sampling steps give higher scores (AGX 30W).
    (d) Dynamic \emph{drift} (Empirical Finding~3): sweet step shifts from 16 (RTX 4070) to 4 (AGX 30W).}
    \label{tab:real-world-results}
    \end{minipage}
    \par
\endgroup

\vspace{-0.8em}
\section{Conclusion}
\label{sec:conclusion}
\vspace{-0.5em}

We present TISED, a task-level decomposition framework for analyzing lossy lightweighting inference optimization in embodied policies.
Beyond inference latency, TISED decomposes optimization into per-chunk speedup, rollout chunk count, action delay, and compute-hardware dependence.
Across policy-intrinsic and execution-aware optimization, which validated in both simulation and real-world UR5e experiments, we identify two paradoxes and one drift:
(1) shorter per-chunk time can yield \emph{slower} per-task time on static tasks;
(2) less precise inference can yield \emph{higher} success on dynamic tasks;
and (3) the per-task sweet spot drifts with the deployment hardware.
More fundamentally, these results elevate closed-loop computation to its own design axis, coupling speed, quality, dynamics, and hardware as joint determinants of task-level performance.

\vspace{-0.6em}
\section{Limitations}
\label{sec:limitations}
\vspace{-6pt}
This work has several limitations.
(1) Our framework treats per-chunk time and chunk count as the dominant degrees of freedom; second-order effects such as scheduling jitter, and OS-level scheduling latency are abstracted away and may matter for more precise analysis.
(2) For execution-aware optimization, we model asynchronous inference as a simplified overlap depth, not as a full server-client implementation; production schedulers may exhibit additional queueing or contention effects.
(3) For the static tasks, we approximate $N$ as a function of action quality alone; other effects such as motion jerk and grasp stability are observed but not formally modeled.
(4) For the dynamic tasks, extremely high-frequency interactions (e.g., table tennis) are out of scope.
(5) Our real-world validation is restricted to a single UR5e platform and two task families; large-scale physical validation is left to future work.

\par\vspace{5pt}\noindent
\begin{minipage}{\linewidth}
\acknowledgments{
This work was supported by the ``U35 Pei Miao Funding Program" of Changping District, Beijing (Grant No. CHPU35202504005).
The corresponding author is Hongyang Jia.
}
\end{minipage}
\par

\bibliography{example}  %
\clearpage
\appendix

\nolinenumbers
\addtocontents{toc}{\protect\setcounter{tocdepth}{2}}
\setcounter{tocdepth}{2}
\tableofcontents
\clearpage

\section{Detailed Analysis of the TISED (Section 3)}
\label{app:analysis}

This section provides a quantitative interpretation of the task-level inference speedup effect decomposition (TISED) introduced in the main paper.
Table~\ref{tab:app-framework-notation} summarizes the notation used in the derivation.

\begin{center}
    \centering
    \footnotesize
    \setlength{\tabcolsep}{3pt}
    \renewcommand{\arraystretch}{1.08}
    \begin{tabular}{p{0.10\linewidth}p{0.36\linewidth}p{0.10\linewidth}p{0.36\linewidth}}
        \toprule
        \rowcolor{black!10}
        Symbol & Meaning & Symbol & Meaning \\
        \midrule
        $H$ & Compute hardware platform. & $\rho_{H}$ & Ratio $T_{\mathrm{inf},H}/T_{\mathrm{act}}$. \\
        $\alpha$ & Policy-intrinsic inference-time ratio, with $\alpha\in[\alpha_{\min},1]$. & $n'$ & Discrete asynchronous overlap depth, with $n'\in\{0,\ldots,n\}$. \\
        $\alpha_{\min}$ & Smallest $\alpha$ whose policy accuracy remains above zero. & & \\
        $T_{\mathrm{inf},H}$ & Baseline inference time on $H$. & $T_{\mathrm{act}}$ & Execution time of one action. \\
        $n$ & Actions in one chunk. & $T_{\mathrm{chunk},H}^{\mathrm{base}}$ & Baseline action-chunk time. \\
        $T_{\mathrm{chunk},H}^{\mathrm{pl/ea}}$ & Optimized per-chunk time. & $N^{\mathrm{pl/ea}}$ & Chunks required for task completion. \\
        $T_{\mathrm{task},H}^{\mathrm{pl/ea}}$ & per-task time. & $D_{\mathrm{act},H}^{\mathrm{pl/ea}}$ & Residual inference-induced waiting time. \\
        $q$ & Action quality. & $\omega_{\mathrm{env}}$ & Environment change rate. \\
        $z_{H}$ & Staleness exposure $D_{\mathrm{act},H}\omega_{\mathrm{env}}$. & $\Phi$ & Task response function. \\
        $\widehat{P}_{\mathrm{succ},H}$ & Modeled dynamic-task success rate. & & \\
        \bottomrule
    \end{tabular}
    \captionsetup{hypcap=false}
    \captionof{table}{Notation used in the quantitative analysis of the TISED framework.}
    \label{tab:app-framework-notation}
\end{center}

\subsection{When Does a ``Sweet Spot'' Exist?}
\label{app:sweetspot-existence}

\subsubsection{Optimization Variables}
\paragraph{Policy-Intrinsic Optimization.}
We analyze the inference-time ratio $\alpha\in[\alpha_{\min},1]$, where smaller $\alpha$ means more aggressive lightweighting.
All derivatives with respect to $\alpha$ below are therefore taken in the direction of \emph{less} aggressive lightweighting.
When reading the same result along the optimization-strength direction used in the plots, the direction is reversed.

\paragraph{Execution-Aware Optimization.}
We analyze the overlap depth $n'\in\{0,\ldots,n\}$, where larger $n^\prime$ means more aggressive overlap between inference and action execution.

\subsubsection{Static Tasks}
\paragraph{Policy-Intrinsic Optimization.}
For static tasks, we have
\begin{equation}
    T_{\mathrm{task},H}^{\mathrm{pl}}(\alpha)
    =
    N^{\mathrm{pl}}(\alpha)T_{\mathrm{chunk},H}^{\mathrm{pl}}(\alpha),
    \qquad
    T_{\mathrm{chunk},H}^{\mathrm{pl}}(\alpha)
    =
    \alpha T_{\mathrm{inf},H} + nT_{\mathrm{act}} .
    \label{eq:app-pl-static-time}
\end{equation}
For this analysis, we assume that increasing $\alpha$ improves action quality and decreases the required chunk count,
i.e., $\mathrm{d}N^{\mathrm{pl}}(\alpha)/\mathrm{d}\alpha<0$,
but each chunk becomes slower.
These two monotone effects act in opposite directions on
$T_{\mathrm{task},H}^{\mathrm{pl}}=N^{\mathrm{pl}}T_{\mathrm{chunk},H}^{\mathrm{pl}}$:
the per-chunk speedup can dominate under mild lightweighting, whereas the growth of $N^{\mathrm{pl}}$ can dominate under aggressive lightweighting.
The derivative of $T_{\rm task,H}^{\rm pl}$ with respect to $\alpha$ can be written as a competition between the cost of slower chunks and the benefit of requiring fewer chunks:
\begin{equation}
  \frac{\mathrm{d}T_{\mathrm{task},H}^{\mathrm{pl}}(\alpha)}{\mathrm{d}\alpha}
  =
  \underbrace{
  N^{\mathrm{pl}}(\alpha)T_{\mathrm{inf},H}
  }_{\text{cost: slower chunks}}
  -
  \underbrace{
  \left[-\frac{\mathrm{d}N^{\mathrm{pl}}(\alpha)}{\mathrm{d}\alpha}\right]
  \left(\alpha T_{\mathrm{inf},H}+nT_{\mathrm{act}}\right)
  }_{\text{benefit: fewer chunks}} .
  \label{eq:app-pl-static-derivative}
\end{equation}
A sufficient condition for an interior minimum is:
\begin{equation}
    \left.
    \frac{\mathrm{d}T_{\mathrm{task},H}^{\mathrm{pl}}(\alpha)}
    {\mathrm{d}\alpha}
    \right|_{\alpha=\alpha_{\min}} < 0,
    \qquad
    \left.
    \frac{\mathrm{d}T_{\mathrm{task},H}^{\mathrm{pl}}(\alpha)}
    {\mathrm{d}\alpha}
    \right|_{\alpha=1} > 0.
    \label{eq:app-pl-interior-condition}
\end{equation}
Otherwise, the shortest $T_{\rm task}$ setting can lie at one endpoint.

\paragraph{Execution-Aware Optimization.}
For execution-aware optimization, the corresponding objective is
\begin{equation}
    T_{\mathrm{task},H}^{\mathrm{ea}}(n')
    =
    N^{\mathrm{ea}}(n')T_{\mathrm{chunk},H}^{\mathrm{ea}}(n'),
    \qquad
    T_{\mathrm{chunk},H}^{\mathrm{ea}}(n')
    =
    T_{\mathrm{inf},H}+nT_{\mathrm{act}}
    -
    \min\!\left\{T_{\mathrm{inf},H},n'T_{\mathrm{act}}\right\}.
    \label{eq:app-ea-cycle}
\end{equation}
Larger $n'$ means more aggressive overlap.
Increasing $n'$ can reduce $T_{\mathrm{chunk},H}^{\mathrm{ea}}$ by hiding more inference behind action execution,
but excessive overlap may increase $N^{\mathrm{ea}}$ because action quality or observation--execution alignment degrades.
Let $\Delta f(n')=f(n'+1)-f(n')$; then
\begin{equation}
    \begin{aligned}
    \Delta T_{\mathrm{task},H}^{\mathrm{ea}}(n')
    &=
    T_{\mathrm{task},H}^{\mathrm{ea}}(n'+1)
    -
    T_{\mathrm{task},H}^{\mathrm{ea}}(n') \\
    &=
    \underbrace{
    \Delta N^{\mathrm{ea}}(n')T_{\mathrm{chunk},H}^{\mathrm{ea}}(n'+1)
    }_{\text{cost: more chunks}}
    -
    \underbrace{
    N^{\mathrm{ea}}(n')\left[-\Delta T_{\mathrm{chunk},H}^{\mathrm{ea}}(n')\right]
    }_{\text{benefit: faster chunks}} .
    \end{aligned}
    \label{eq:app-ea-difference}
\end{equation}
For $n\geq2$, a sufficient condition for an interior discrete minimum is:
\begin{equation}
    \Delta T_{\mathrm{task},H}^{\mathrm{ea}}(0)<0,
    \qquad
    \Delta T_{\mathrm{task},H}^{\mathrm{ea}}(n-1)>0.
    \label{eq:app-ea-interior-condition}
\end{equation}
Otherwise, the shortest $T_{\rm task}$ setting can lie at one endpoint.

\subsubsection{Dynamic Tasks}

\paragraph{Policy-Intrinsic Optimization.}
We model the policy-intrinsic dynamic-task success rate as
\begin{equation}
    \widehat{P}_{\mathrm{succ},H}^{\mathrm{pl}}(\alpha)
    \mathrel{=}
    \Phi\!\left(q^{\mathrm{pl}}(\alpha),z_{H}^{\mathrm{pl}}(\alpha)\right),
    \qquad
    z_{H}^{\mathrm{pl}}(\alpha)
    =
    D_{\mathrm{act},H}^{\mathrm{pl}}(\alpha)\omega_{\mathrm{env}}
    =
    \alpha T_{\mathrm{inf},H}\omega_{\mathrm{env}} .
    \label{eq:app-pl-dynamic-surrogate}
\end{equation}
Here $\Phi$ is assumed to be monotone in the two arguments:
\[
\Phi_q=\frac{\partial \Phi}{\partial q}\geq 0,
\qquad
\Phi_z=\frac{\partial \Phi}{\partial z}\leq 0,
\]
meaning that better action quality increases the modeled success rate, while larger staleness exposure decreases it.
Increasing $\alpha$ usually improves action quality but also increases action delay:
$\mathrm{d}q^{\mathrm{pl}}(\alpha)/\mathrm{d}\alpha\geq 0$ and
$\mathrm{d}z_{H}^{\mathrm{pl}}(\alpha)/\mathrm{d}\alpha=T_{\mathrm{inf},H}\omega_{\mathrm{env}}>0$.
Thus, differentiating the surrogate response gives
\begin{equation}
    \frac{\mathrm{d}\widehat{P}_{\mathrm{succ},H}^{\mathrm{pl}}(\alpha)}
    {\mathrm{d}\alpha}
    \mathrel{=}
    \underbrace{
    \Phi_q
    \frac{\mathrm{d}q^{\mathrm{pl}}(\alpha)}{\mathrm{d}\alpha}
    }_{\text{benefit: better action quality}}
    -
    \underbrace{
    \left[-\Phi_z\right]T_{\mathrm{inf},H}\omega_{\mathrm{env}}
    }_{\text{cost: larger action delay}} .
    \label{eq:app-pl-dynamic-derivative-existence}
\end{equation}
A sufficient condition for an interior dynamic-task maximum is:
\begin{equation}
    \left.
    \frac{\mathrm{d}\widehat{P}_{\mathrm{succ},H}^{\mathrm{pl}}}{\mathrm{d}\alpha}
    \right|_{\alpha=\alpha_{\min}} > 0,
    \qquad
    \left.
    \frac{\mathrm{d}\widehat{P}_{\mathrm{succ},H}^{\mathrm{pl}}}{\mathrm{d}\alpha}
    \right|_{\alpha=1} < 0 .
    \label{eq:app-pl-dynamic-interior-condition}
\end{equation}
Otherwise, the best $\widehat{P}_{\rm succ}$ setting can lie at one of the endpoints.

\paragraph{Execution-Aware Optimization.}
The same response model becomes
\begin{equation}
    \widehat{P}_{\mathrm{succ},H}^{\mathrm{ea}}(n')
    \mathrel{=}
    \Phi\!\left(q^{\mathrm{ea}}(n'),z_{H}^{\mathrm{ea}}(n')\right),
    \qquad
    z_{H}^{\mathrm{ea}}(n')
    =
    \max\{0,T_{\mathrm{inf},H}-n'T_{\mathrm{act}}\}\omega_{\mathrm{env}} .
    \label{eq:app-ea-dynamic-surrogate}
\end{equation}
The surrogate represents observation--execution misalignment through $q^{\mathrm{ea}}$ and environmental drift during residual waiting through $z_H^{\mathrm{ea}}$.
Increasing $n'$ means more aggressive execution-aware optimization.
It can reduce residual waiting by hiding inference behind action execution, but it may also degrade action quality or observation--execution alignment.
Let $\Delta f(n')=f(n'+1)-f(n')$.
Then $\Delta z_{H}^{\mathrm{ea}}(n')\leq 0$ before overlap saturation, while $\Delta q^{\mathrm{ea}}(n')$ can be negative when deeper overlap hurts action quality.
A first-order finite-difference approximation gives
\begin{equation}
    \Delta \widehat{P}_{\mathrm{succ},H}^{\mathrm{ea}}(n')
    \approx
    \underbrace{
    \left[-\Phi_z\right]
    \left[-\Delta z_{H}^{\mathrm{ea}}(n')\right]
    }_{\text{benefit: smaller action delay}}
    -
    \underbrace{
    \Phi_q
    \left[-\Delta q^{\mathrm{ea}}(n')\right]
    }_{\text{cost: degraded action quality}} .
    \label{eq:app-ea-dynamic-difference-existence}
\end{equation}
For $n\geq2$, a sufficient condition, using exact differences, for an interior discrete dynamic-task maximum is
\begin{equation}
    \Delta \widehat{P}_{\mathrm{succ},H}^{\mathrm{ea}}(0)>0,
    \qquad
    \Delta \widehat{P}_{\mathrm{succ},H}^{\mathrm{ea}}(n-1)<0.
    \label{eq:app-ea-dynamic-interior-condition}
\end{equation}
Otherwise, the best $\widehat{P}_{\rm succ}$ setting can lie at one endpoint.

\subsection{How Does the ``Sweet Spot'' Move with Hardware?}
\label{app:sweetspot-rho-drift}

\subsubsection{Static Tasks: Take Execution-Aware Optimization as an Example}
{
We assume
$N^{\mathrm{ea}}(n^\prime+1)\geq N^{\mathrm{ea}}(n^\prime)$.
The hardware-dependent sweet spot is
\begin{equation}
    (n^\prime)_{H}^\star
    \in
    \arg\min_{n^\prime\in\{0,\ldots,n\}}
    N^{\mathrm{ea}}(n^\prime)\left(\rho_{H}+n-\min(\rho_{H},n^\prime)\right).
    \label{eq:app-ea-finite-argmin}
\end{equation}

For a local interpretation, we relax the integer overlap depth $n'$ to a continuous variable $x\in[0,n]$, keeping $N^{\mathrm{ea}}(x)$ fixed across hardware.
Suppose reference hardware $H_0$ has an interior sweet spot $x_0^\star<\rho_0$. Then
\begin{equation}
    T_{\mathrm{task}}^{\mathrm{ea}}(x;\rho)
    =
    T_{\mathrm{act}}N^{\mathrm{ea}}(x)(\rho+n-x).
    \label{eq:app-ea-local-objective}
\end{equation}
The complete derivative with respect to $x$ and the reference sweet-spot condition are
\begin{equation}
    \begin{aligned}
    \frac{\partial T_{\mathrm{task}}^{\mathrm{ea}}}{\partial x}
    &=
    T_{\mathrm{act}}
    \left[
    \frac{\mathrm{d}N^{\mathrm{ea}}(x)}{\mathrm{d}x}
    (\rho+n-x)
    -
    N^{\mathrm{ea}}(x)
    \right],\\
    \left.
    \frac{\partial T_{\mathrm{task}}^{\mathrm{ea}}}{\partial x}
    \right|_{x=x_0^\star,\rho=\rho_0}
    &=
    \left.
    T_{\mathrm{act}}
    \left[
    \frac{\mathrm{d}N^{\mathrm{ea}}(x)}{\mathrm{d}x}
    (\rho_0+n-x)
    -
    N^{\mathrm{ea}}(x)
    \right]
    \right|_{x=x_0^\star}
    =
    0.
    \end{aligned}
    \label{eq:app-ea-full-derivative}
\end{equation}
Evaluating the slope at the same reference point on hardware $\rho_1>x_0^\star$ gives
\begin{equation}
    \left.
    \frac{\partial T_{\mathrm{task}}^{\mathrm{ea}}}{\partial x}
    \right|_{x=x_0^\star,\rho=\rho_1}
    =
    T_{\mathrm{act}}(\rho_1-\rho_0)
    \left.
    \frac{\mathrm{d}N^{\mathrm{ea}}(x)}{\mathrm{d}x}
    \right|_{x=x_0^\star}.
    \label{eq:app-ea-local-drift}
\end{equation}

Slower hardware with $\rho_1>\rho_0$ increases the slope at the reference sweet spot, so the sweet spot tends to shift toward a smaller overlap depth.

If the new hardware places the setting in the saturated branch, i.e., $x\geq\rho_1$, the per-chunk time becomes
$T_{\mathrm{chunk}}^{\mathrm{ea}}=nT_{\mathrm{act}}$ and no longer decreases with deeper overlap.
Under $\frac{\mathrm{d}N^{\mathrm{ea}}}{\mathrm{d}x}\geq0$, increasing $x$ in this branch cannot reduce $T_{\mathrm{task}}^{\mathrm{ea}}$.
}

\subsubsection{Dynamic Tasks: Take Policy-Intrinsic Optimization as an Example}
{
For the measured policy-intrinsic sweep,
\begin{equation}
    \alpha_{H}^\star
    \in
    \arg\max_{\alpha_i\in\mathcal{A}}
    \Phi\!\left(q_i^{\mathrm{pl}},
    \alpha_i\rho_{H}T_{\mathrm{act}}\omega_{\mathrm{env}}\right),
    \label{eq:app-pl-dynamic-argmax}
\end{equation}
where $q_i^{\mathrm{pl}}$ denotes the  action quality at setting $\alpha_i$.
For a local interpretation, we keep the optimization settings and their action quality fixed across hardware. Suppose reference hardware $H_0$ with $\rho_0$ has an interior sweet spot $\alpha_0^\star$. Then
\begin{equation}
    \widehat{P}_{\mathrm{succ}}^{\mathrm{pl}}(\alpha;\rho)
    \mathrel{=}
    \Phi\!\left(q^{\mathrm{pl}}(\alpha),
    \alpha\rho T_{\mathrm{act}}\omega_{\mathrm{env}}\right),
    \label{eq:app-pl-dynamic-local-objective}
\end{equation}
Let $z^{\mathrm{pl}}(\alpha,\rho)=\alpha\rho T_{\mathrm{act}}\omega_{\mathrm{env}}$.
The complete derivative with respect to $\alpha$ is
\begin{equation}
    \frac{\partial \widehat{P}_{\mathrm{succ}}^{\mathrm{pl}}}{\partial \alpha}
    =
    \Phi_q\!\left(q^{\mathrm{pl}}(\alpha),z^{\mathrm{pl}}(\alpha,\rho)\right)
    \frac{\mathrm{d}q^{\mathrm{pl}}(\alpha)}{\mathrm{d}\alpha}
    +
    \Phi_z\!\left(q^{\mathrm{pl}}(\alpha),z^{\mathrm{pl}}(\alpha,\rho)\right)
    \rho T_{\mathrm{act}}\omega_{\mathrm{env}},
    \label{eq:app-pl-dynamic-full-derivative}
\end{equation}
where $\Phi_q$ and $\Phi_z$ denote the partial derivatives of $\Phi$ with respect to action quality and staleness exposure, respectively.
The reference sweet spot therefore satisfies
\begin{equation}
    \left[
    \Phi_q
    \frac{\mathrm{d}q^{\mathrm{pl}}}{\mathrm{d}\alpha}
    +
    \Phi_z
    \rho_0 T_{\mathrm{act}}\omega_{\mathrm{env}}
    \right]_{\alpha=\alpha_0^\star,\rho=\rho_0}
    =
    0.
    \label{eq:app-pl-dynamic-reference}
\end{equation}
Ignoring higher-order sensitivity changes of $\Phi$, the local slope after changing hardware to $\rho_1$ becomes
\begin{equation}
    \left.
    \frac{\partial \widehat{P}_{\mathrm{succ}}^{\mathrm{pl}}}{\partial\alpha}
    \right|_{\alpha=\alpha_0^\star,\rho=\rho_1}
    \mathrel{\approx}
    (\rho_1-\rho_0)T_{\mathrm{act}}\omega_{\mathrm{env}}
    \Phi_z.
\label{eq:app-pl-dynamic-drift}
\end{equation}

Here $z=\alpha\rho T_{\mathrm{act}}\omega_{\mathrm{env}}$ denotes staleness exposure, and $\Phi_z\leq0$ means that larger staleness lowers success.
Under this approximation, slower hardware with $\rho_1>\rho_0$ makes the slope at the reference sweet spot negative when $\Phi_z<0$, so the sweet spot tends to shift toward smaller $\alpha$.
}

\subsection{Assumptions and Limitations}
\label{app:analysis-limitations}

The analysis above makes several idealized assumptions.
First, quantities such as $N^{\mathrm{pl}}(\alpha)$ and $N^{\mathrm{ea}}(n^\prime)$ are specified only through their qualitative trends.
Their concrete functional forms are task-dependent and must be obtained from empirical measurement.
Second, $\Phi$ is used only as an abstract response function that describes how dynamic-task success depends on action quality and action delay.
We do not assume that $\Phi$ has a fixed parametric form or represents a calibrated physical law.

\section{Simulation Experiment Details (Section 4.2, 4.3, 4.4, 4.5)}
\label{app:simulation}

This section describes how we design the optimization methods and choose the task regimes.

\subsection{Optimization Method Settings}
\label{app:simulation-optimization}

We summarize the policy models used in the main paper and their corresponding static and dynamic evaluation configurations in Tables~\ref{tab:app-action-exec-config} and~\ref{tab:app-dynamic-config}.

\begin{table}[!htbp]
    \centering
    \scriptsize
    \setlength{\tabcolsep}{2.1pt}
    \renewcommand{\arraystretch}{1.06}
    \begin{tabular}{lp{0.30\linewidth}ccccc}
        \toprule
        \rowcolor{black!10}
        Model & Task and optimization method &
        \begin{tabular}[c]{@{}c@{}}Default\\action-\\generation\\steps\end{tabular} &
        $n$ & \begin{tabular}[c]{@{}c@{}}Action\\freq. (Hz)\end{tabular} &
        $T_{\mathrm{exec}}$ (ms) & $nT_{\mathrm{exec}}$ (ms) \\
        \midrule
        Cosmos-Policy & LIBERO \& Asynchronous Inference; RoboCasa \& Quantization & 1 & 16 & 20 & 50.0 & 800.0 \\
        LingBot-VA & RoboTwin 2.0 \& Pruning & 10 & 2 & 16.7 & 60.0 & 120.0 \\
        $\pi_{0.5}$ & LIBERO \& Asynchronous Inference & 10 & 5 & 30 & 33.3 & 166.7 \\
        \bottomrule
    \end{tabular}
    \caption{Simulation static-task configurations.}
    \label{tab:app-action-exec-config}
\end{table}

\begin{table}[!htbp]
    \centering
    \small
    \setlength{\tabcolsep}{5pt}
    \renewcommand{\arraystretch}{1.06}
    \begin{tabular}{llr}
        \toprule
        \rowcolor{black!10}
        Model & Task and optimization method & Default sampling steps \\
        \midrule
        MLP-Mixer & Kinetix \& flow-style sampling-step reduction & 5 \\
        DynamicVLA & DOM-CR \& flow-matching sampling-step reduction & 10 \\
        \bottomrule
    \end{tabular}
    \caption{Simulation dynamic-task configurations.}
    \label{tab:app-dynamic-config}
\end{table}

\paragraph{Quantization.}
At the operator level, we replace the FP16 linear layers in Transformer-based policies with integer linear operators, including int4-weight/int4-activation (W4A4) and int8-weight/int8-activation (W8A8) variants.
The additional quantization and dequantization operators are fused into the surrounding operators to reduce inference overhead.
We do not apply extra transformations such as smoothing~\cite{xiao2023smoothquant} or rotation~\cite{ashkboos2024quarotoutlierfree4bitinference}.
To obtain fine-grained operating points and expose potential sweet spots, we construct a progressive tier-10 quantization sweep as in~\cite{kang2026win}; the Cosmos-Policy quantization configurations are listed in Table~\ref{tab:app-quant-config}.

\begin{figure}[t]
\centering
\includegraphics[width=\linewidth]{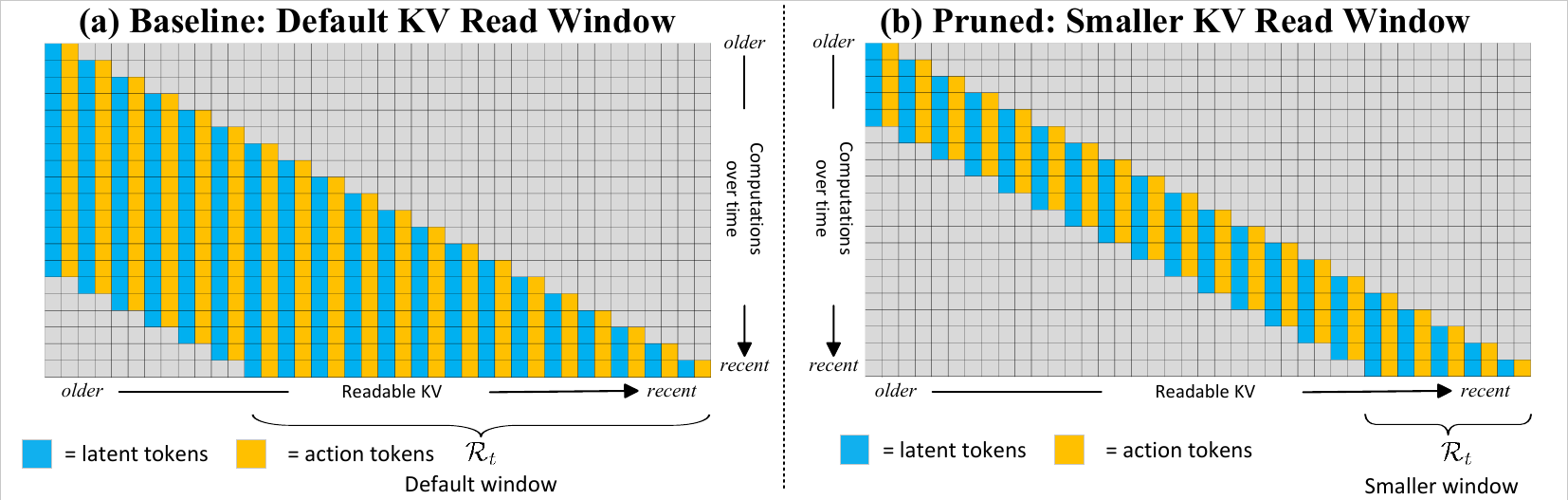}
\caption{KV-cache read-window pruning for LingBot-VA. The baseline attends to all valid visual-latent and action KV slots retained by the native attention window. The pruned setting further restricts attention to the most recent readable cache windows, while the cache-writing path remains unchanged.}
\label{fig:app-pr-cache}
\end{figure}

\newcommand{\qconfigcell}[1]{\begin{tabular}[t]{@{}l@{}}#1\end{tabular}}
\newcommand{\qconfigrowsep}{\arrayrulecolor{black!18}\specialrule{0.25pt}{1pt}{1pt}\arrayrulecolor{black}}
\begin{table}[!htbp]
    \centering
    \scriptsize
    \setlength{\tabcolsep}{2.2pt}
    \renewcommand{\arraystretch}{1.08}
    \begin{tabular}{p{0.09\linewidth}p{0.12\linewidth}p{0.36\linewidth}p{0.32\linewidth}}
        \toprule
        \rowcolor{black!10}
        Main label & Speedup vs. FP16 & W4A4 sites & Remaining higher-precision sites \\
        \midrule
        fp16 & $1.00\times$ & None & All candidate sites remain FP16. \\
        \qconfigrowsep
        w8a8 & $1.33\times$ & None & All candidate sites use W8A8. \\
        \qconfigrowsep
        w4-t1 & $1.36\times$ & \qconfigcell{\texttt{ca.out@late}} & \qconfigcell{W8A8:\\\texttt{ca.out@early/mid}\\\texttt{sa.*@all}\\\texttt{ca.q/k/v@all}\\\texttt{mlp.l1/l2@all}} \\
        \qconfigrowsep
        w4-t2 & $1.41\times$ & \qconfigcell{\texttt{ca.out@late}\\\texttt{mlp.l1@early}} & \qconfigcell{W8A8:\\\texttt{ca.out@early/mid}\\\texttt{mlp.l1@mid/late}\\\texttt{sa.*@all}\\\texttt{ca.q/k/v@all}\\\texttt{mlp.l2@all}} \\
        \qconfigrowsep
        w4-t3 & $1.42\times$ & \qconfigcell{\texttt{ca.out@early/late}\\\texttt{mlp.l1@early}} & \qconfigcell{W8A8:\\\texttt{ca.out@mid}\\\texttt{mlp.l1@mid/late}\\\texttt{sa.*@all}\\\texttt{ca.q/k/v@all}\\\texttt{mlp.l2@all}} \\
        \qconfigrowsep
        w4-t4 & $1.47\times$ & \qconfigcell{\texttt{ca.out@early/late}\\\texttt{mlp.l1@early/late}} & \qconfigcell{W8A8:\\\texttt{ca.out@mid}\\\texttt{mlp.l1@mid}\\\texttt{sa.*@all}\\\texttt{ca.q/k/v@all}\\\texttt{mlp.l2@all}} \\
        \qconfigrowsep
        w4-t5 & $1.53\times$ & \qconfigcell{\texttt{ca.out@early/late}\\\texttt{mlp.l1@all}} & \qconfigcell{W8A8:\\\texttt{ca.out@mid}\\\texttt{sa.*@all}\\\texttt{ca.q/k/v@all}\\\texttt{mlp.l2@all}} \\
        \qconfigrowsep
        w4-t6 & $1.55\times$ & \qconfigcell{\texttt{ca.out@all}\\\texttt{mlp.l1@all}} & \qconfigcell{W8A8:\\\texttt{sa.*@all}\\\texttt{ca.q/k/v@all}\\\texttt{mlp.l2@all}} \\
        \qconfigrowsep
        w4-t7 & $1.67\times$ & \qconfigcell{\texttt{ca.out@all}\\\texttt{mlp.l1/l2@all}} & \qconfigcell{W8A8:\\\texttt{sa.*@all}\\\texttt{ca.q/k/v@all}} \\
        \qconfigrowsep
        w4-t8 & $1.88\times$ & \qconfigcell{\texttt{ca.out@all}\\\texttt{mlp.l1/l2@all}\\\texttt{sa.q/k/out@all}} & \qconfigcell{W8A8:\\\texttt{sa.v@all}\\\texttt{ca.q/k/v@all}} \\
        \qconfigrowsep
        w4-t9 & $2.05\times$ & \qconfigcell{\texttt{sa.q/k/v/out@all}\\\texttt{ca.q/out@all}\\\texttt{mlp.l1/l2@all}} & \qconfigcell{W8A8:\\\texttt{ca.k/v@all}} \\
        \qconfigrowsep
        w4-t10 & $2.09\times$ & \qconfigcell{\texttt{sa.q/k/v/out@all}\\\texttt{ca.q/k/v/out@all}\\\texttt{mlp.l1/l2@all}} & None; all candidate sites use W4A4. \\
        \bottomrule
    \end{tabular}
    \caption{Mapping between the quantization shorthand and the progressive tier-10 Cosmos-Policy quantization settings. Speedup is measured from LIBERO 5-step latency profiling relative to all-FP16. The sweep starts from all-W8A8 and progressively converts selected DiT attention/MLP linear sites to W4A4. Here, \texttt{early/mid/late} denote DiT blocks \texttt{B0--B9/B10--B19/B20--B27}; \texttt{sa}, \texttt{ca}, and \texttt{mlp} denote self-attention, cross-attention, and feed-forward MLP sites; \texttt{q/k/v/out} denote attention projections, and \texttt{l1/l2} denote the two MLP linear layers.}
    \label{tab:app-quant-config}
\end{table}
\paragraph{Pruning.}
For the pruning sweep, we use LingBot-VA~\cite{li2026causal} on RoboTwin 2.0 tasks and implement a KV-cache pruning method. In the default RoboTwin 2.0 setting, \texttt{attn\_window}=72, which allocates up to 36 historical visual-latent chunks and 36 historical action chunks in the baseline cache. Our pruning further shortens the readable window inside this native cache. Let $\mathcal{C}^{\mathrm{vis}}_{<t}$ and $\mathcal{C}^{\mathrm{act}}_{<t}$ denote the valid visual-latent and action KV slots retained by the native attention window before step $t$.
The pruned attention reads only
\begin{equation}
\mathcal{R}_t
=
\operatorname{Recent}_{W_{\mathrm{lat}}}
\left(\mathcal{C}^{\mathrm{vis}}_{<t}\right)
\cup
\operatorname{Recent}_{W_{\mathrm{act}}}
\left(\mathcal{C}^{\mathrm{act}}_{<t}\right),
\qquad
h_t=\operatorname{Attn}\!\left(q_t,K_{\mathcal{R}_t},V_{\mathcal{R}_t}\right),
\label{eq:kv-read-window-pruning}
\end{equation}
where $\operatorname{Recent}_{W}(\cdot)$ keeps the most recent $W$ slots of the corresponding cache stream.
Figure~\ref{fig:app-pr-cache} illustrates the difference between the native-window baseline and our read-window pruning.
The pruned setting keeps cache writing unchanged but exposes only the most recent $W_{\mathrm{lat}}$ and $W_{\mathrm{act}}$ entries to attention.

Table~\ref{tab:app-lingbot-prune-config} lists the operating points used in the sweep.
To make the pruning strength directly comparable across settings, we first report the total readable KV budget retained relative to the baseline.
This normalized value decreases monotonically from \texttt{base} to \texttt{p4}, so smaller values indicate more aggressive lightweighting.
The KV-budget columns then report the maximum number of readable historical tokens for the visual-latent and action streams.
The chunk columns convert these token budgets using the RoboTwin 2.0 configuration. Each visual-latent chunk contains 240 KV tokens, and each action chunk contains 32 KV tokens.
For example, $1024/256$ means that each action-generation step can read at most the most recent 1024 visual-latent KV entries and 256 action KV entries, corresponding to roughly 4.3 visual chunks and 8 action chunks.

\begin{table}[t]
\centering
\scriptsize
\resizebox{\linewidth}{!}{%
\begin{tabular}{cccccc}
\toprule
\rowcolor{black!10}
Config &
\begin{tabular}[c]{@{}c@{}}Readable KV Compression Ratio\end{tabular} &
Visual KV budget &
Action KV budget &
Visual hist. chunks &
Action hist. chunks \\
\midrule
base & 100.0\% & 8640 & 1152 & 36.0 & 36.0 \\
p1 & 13.1\% & 1024 & 256 & 4.3 & 8.0 \\
p2 & 6.5\% & 512 & 128 & 2.1 & 4.0 \\
p3 & 3.3\% & 256 & 64 & 1.1 & 2.0 \\
p4 & 1.6\% & 128 & 32 & 0.5 & 1.0 \\
\bottomrule
\end{tabular}
}
\caption{LingBot-VA KV-cache read-window configurations. The ``readable KV Compression Ratio'' column reports the combined readable visual and action KV budget as a percentage of the baseline, so smaller values indicate more aggressive lightweighting. The baseline budget is determined by the native attention window, and pruning reduces only the readable KV budget inside that window.}
\label{tab:app-lingbot-prune-config}
\end{table}

\paragraph{Action-Generation Step-Count Reduction.}
We reduce the number of action-generation steps used to generate each action chunk.
We then evaluate a progressive step sweep to characterize the trade-off between computing speed and action quality.

\paragraph{Asynchronous Inference.}
We adopt the simplified asynchronous-inference model from VLASH~\cite{tang2025vlash}.
The overlap depth between inference and execution is controlled by $n^\prime$: at simulation step $t$, the policy receives the stale observation recorded at step $t-n^\prime$.
This emulates delayed machine-vision input due to overlapping policy inference and action execution.

\subsection{Task Settings}
\label{app:simulation-tasks}
We group the simulation benchmarks according to the task regimes defined in the main paper: LIBERO, RoboCasa, and the RoboTwin 2.0 tasks are used as static-task benchmarks, while Kinetix and DOM-CR are used as dynamic-task benchmarks.

\paragraph{LIBERO.}
LIBERO~\cite{liu2023libero} is a tabletop manipulation benchmark suite designed to evaluate policy generalization and knowledge transfer across object, spatial, goal, and long-horizon task variations.
We use it as a controlled static-task setting because the task-relevant object and goal states are assumed to remain unchanged during policy inference.

\paragraph{RoboCasa.}
RoboCasa~\cite{robocasa2024} provides large-scale simulated household manipulation tasks with diverse kitchen layouts, objects, and task instances.
Despite the static-scene assumption, RoboCasa tasks are typically more challenging than those in LIBERO.

\paragraph{RoboTwin 2.0.}
We evaluate LingBot-VA on RoboTwin 2.0~\cite{chen2025robotwin2}, a SAPIEN-based benchmark for bimanual robotic manipulation that provides diverse tasks, multi-camera observations, and standardized evaluation protocols.
In our evaluation setting, these tasks are categorized as static manipulation tasks because no external motion is imposed on task-relevant objects during policy inference.

\paragraph{Kinetix.}
Kinetix~\cite{matthews2024kinetix} is a 2D physics-based control benchmark with procedurally designed tasks involving contact-rich interaction, object manipulation, locomotion, balance, and timing-sensitive control.
The policy is an iterative, flow-style action generator with an MLP-Mixer architecture~\cite {tang2025vlash}.
We evaluate both zero-delay inference, denoted as \texttt{inf-hw} in the main text, and hardware-induced delay.
We use two latency-simulation protocols with the following common settings.
The native physics timestep is $\Delta=1/60\,\mathrm{s}\approx16.67\,\mathrm{ms}$; each control interval spans 2 such steps, so $T_{\mathrm{act}}=2\Delta\approx33.33\,\mathrm{ms}$.
Each policy call predicts 8 actions; the policy replans from the latest observation after every 4 control intervals and retains the prediction's final 4 actions as the next window's fallback candidates.
The nominal replanning period is therefore $4T_{\mathrm{act}}=8/60\,\mathrm{s}\approx133.33\,\mathrm{ms}$ in virtual simulation time.

\noindent\textit{Coarse-grained simulation.}
We vary the number of sampling steps in $\{1,3,5\}$ to trade off zero-delay action quality against inference latency in the coarse-grained evaluation (Figure~\ref{fig:obs2}(a)).
We map profiled inference latency $\tau$ to an integer control-step delay $d=\mathrm{round}(\tau/T_{\mathrm{act}})$ using nearest-even rounding, giving effective delay $dT_{\mathrm{act}}$.
Consequently, different profiled latencies can fall into the same simulator delay bin.

\begin{table}[!htbp]
    \centering
    \scriptsize
    \setlength{\tabcolsep}{3.2pt}
    \renewcommand{\arraystretch}{1.06}
    \begin{tabular}{lrrrr}
        \toprule
        \rowcolor{black!10}
        Execution setting & Steps & $\tau$ (ms) & $d$ & Effective delay (ms) \\
        \midrule
        6000 Ada & 1 & 1.80 & 0 & 0.00 \\
        6000 Ada & 3 & 3.81 & 0 & 0.00 \\
        6000 Ada & 5 & 9.82 & 0 & 0.00 \\
        \midrule
        AGX 30W & 1 & 7.78 & 0 & 0.00 \\
        AGX 30W & 3 & 21.44 & 1 & 33.33 \\
        AGX 30W & 5 & 35.80 & 1 & 33.33 \\
        \midrule
        AGX 15W & 1 & 7.78 & 0 & 0.00 \\
        AGX 15W & 3 & 33.75 & 1 & 33.33 \\
        AGX 15W & 5 & 54.63 & 2 & 66.67 \\
        \bottomrule
    \end{tabular}
    \caption{Kinetix coarse-grained latency-to-delay configuration for Figures~\ref{fig:obs2}(a) and~\ref{fig:obs3}(b).}
    \label{tab:app-kinetix-delay-config}
\end{table}

\begin{figure}[!htbp]
    \centering
    \includegraphics[width=\linewidth]{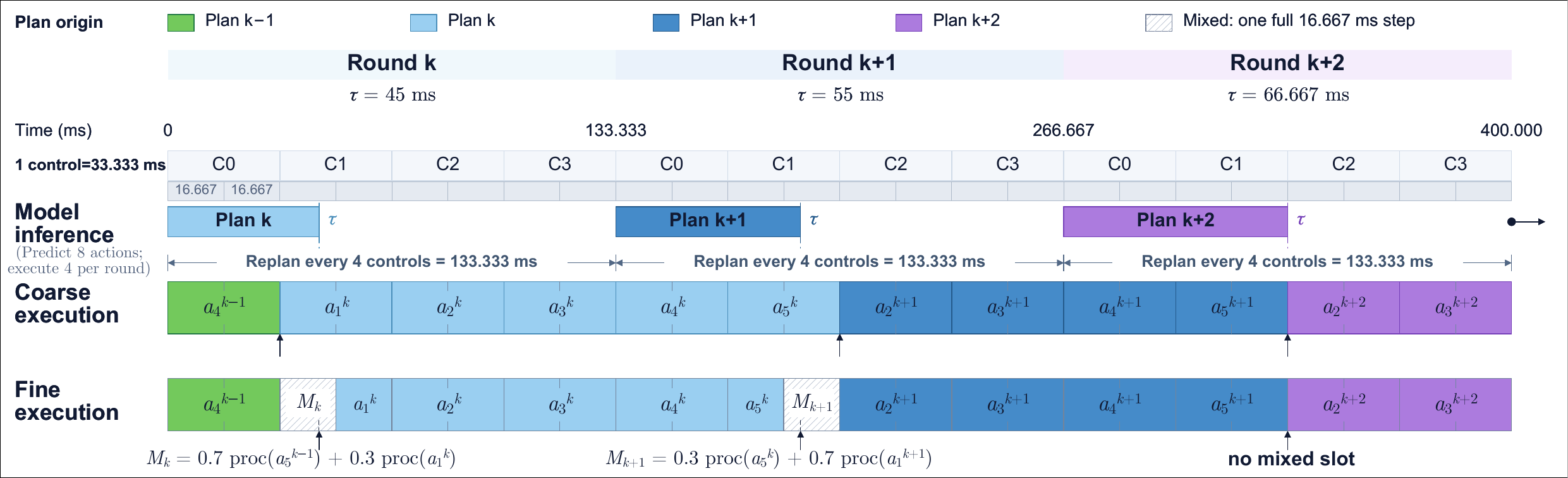}
    \caption{Visualization of coarse- and fine-grained simulation in Kinetix.}
    \label{fig:kinetix-latency-replay}
\end{figure}

\noindent\textit{Fine-grained simulation.}
{
To simulate inference delays at a finer temporal resolution without changing the simulator's update logic, we use a fine-grained protocol for the predictive TISED case study in Section~\ref{sec:predictive-case} (Figure~\ref{fig:kinetix-latency-replay}).
Each replanning window contains 8 physics steps, indexed by $j\in\{0,\ldots,7\}$.
We approximate a switch from old to new commands within a physics step using a single time-weighted mixture.
For step $j$, covering $[j\Delta,(j+1)\Delta]$, $w_j$ is the fraction of the step before the new prediction becomes available at time $\tau$.
The applied command $\mathbf{u}_j$ combines the old and new actuator commands, $\mathbf{u}_j^{\mathrm{old}}$ and $\mathbf{u}_j^{\mathrm{new}}$, according to this fraction:
\begin{equation}
    w_j=\operatorname{clip}\!\left(\frac{\tau}{\Delta}-j,0,1\right),
    \qquad
    \mathbf{u}_j=w_j\mathbf{u}_j^{\mathrm{old}}+(1-w_j)\mathbf{u}_j^{\mathrm{new}}.
    \label{eq:kinetix-fine-command}
\end{equation}
At round $r$, the candidates for control position $i=\lfloor j/2\rfloor$ come from $a^{(r-1)}_{4+i}$ in the previous plan's retained tail and $a^{(r)}_i$ in the current plan.
Both candidate actions are converted into actuator commands before mixing.
Each physics step uses an old, mixed, or new command according to $w_j$.
\par}

\noindent\textit{Predictive TISED details.}
{
The case study in Section~\ref{sec:predictive-case} evaluates \textit{Grasp}, \textit{Half-cheetah}, and \textit{Walker} using the fine-grained simulation above, with $s\in\{1,2,3,4,5\}$ sampling steps and $1{,}024$ paired stochastic seeds per condition at a fixed initial state for each task.
We choose these tasks to represent 3 response patterns: \textit{Grasp} is mainly sensitive to delay, \textit{Half-cheetah} to lightweighting, and \textit{Walker} to both.
We measure inference latency $T_{\mathrm{inf},H}(s)$ for each hardware setting $H$ (Table~\ref{tab:app-predictive-latencies}) and define
\begin{equation}
    \alpha_H(s)=\frac{T_{\mathrm{inf},H}(s)}{T_{\mathrm{inf},H}(5)},
    \qquad
    D_H(\alpha_H)=T_{\mathrm{inf},H}(5)\alpha_H.
    \label{eq:app-predictive-hardware}
\end{equation}
For each hardware platform, the slope equals its measured inference latency at $s=5$.
In the normalized notation of Section~\ref{sec:predictive-case}, $f_{\mathrm{HW}}(\alpha_H)=\alpha_H$.
\par}

\begin{table}[!htbp]
    \centering

    \small
    \setlength{\tabcolsep}{5pt}
    \renewcommand{\arraystretch}{1.08}
    \begin{tabular}{lrrrrr}
        \toprule
        \rowcolor{black!10}
        Hardware & $s=1$ & $s=2$ & $s=3$ & $s=4$ & $s=5$ \\
        \midrule
\textit{RTX 6000 Ada} & 1.37 & 2.60 & 4.04 & 5.13 & 6.90 \\
\textit{RTX 3090} & 1.85 & 3.54 & 5.23 & 7.07 & 8.81 \\
\textit{AGX Orin 30W} & 7.46 & 14.34 & 21.94 & 29.53 & 36.74 \\
\textit{AGX Orin 15W} & 11.35 & 22.74 & 32.68 & 46.16 & 57.63 \\
        \bottomrule
    \end{tabular}
    \caption{Inference latencies in milliseconds used for the predictive TISED case study in Figure~\ref{fig:predictive-case}.}
    \label{tab:app-predictive-latencies}
\end{table}

{
For calibration, we vary $s$ at zero delay to measure the quality response, and fix $s=5$ while varying delay over 9 equally spaced values from $0$ to $200/3\,\mathrm{ms}$ to measure the delay response.
We select constant, logistic, or linear response families for each task and fit their parameters to calibration success counts by binomial maximum likelihood.
We first fit $\mathrm{acc}_s g(\alpha_{\mathrm{Ada}}(s))$ to the zero-delay results, where $\mathrm{acc}_s$ is the fitted success rate at $s=5$.
We then hold $\mathrm{acc}_s$ fixed and fit $\mathrm{acc}_s h(D)$ to the delay observations, with $h(0)=1$.
\par}

\begin{table}[!htbp]
    \centering

    \small
    \setlength{\tabcolsep}{5pt}
    \renewcommand{\arraystretch}{1.75}
    \begin{tabular}{lccl}
        \toprule
        \rowcolor{black!10}
        Task & $\mathrm{acc}_{5}$ & Response & Fitted expression \\
        \midrule
        \textit{Grasp} & $0.97$ & $g(\alpha)$ & $1$ \\
         & & $h(D)$ & $\displaystyle 1-0.77\frac{\exp(0.26D)-1}{\exp(0.26D)+153.27}$ \\
        \midrule
        \textit{Half-cheetah} & $0.87$ & $g(\alpha)$ & $\displaystyle \frac{1.00}{1+\exp[-8.54(\alpha+3.58\times10^{-4})]}$ \\
         & & $h(D)$ & $1-(8.74\times10^{-4})D$ \\
        \midrule
        \textit{Walker} & $0.75$ & $g(\alpha)$ & $\displaystyle \frac{1.00}{1+\exp[-24.35(\alpha-0.29)]}$ \\
         & & $h(D)$ & $1-(1.01\times10^{-2})D$ \\
        \bottomrule
    \end{tabular}
    \caption{Fitted responses for the predictive TISED case study in Figure~\ref{fig:predictive-case}.}
    \label{tab:app-predictive-fits}
\end{table}

\begin{table}[!htbp]
    \centering

    \small
    \setlength{\tabcolsep}{4pt}
    \renewcommand{\arraystretch}{1.08}
    \begin{subtable}[t]{0.48\linewidth}
        \centering
        \caption{Zero-delay policy response ($D=0$).}
        \label{tab:app-predictive-quality-data}
        \begin{tabular*}{\linewidth}{@{\extracolsep{\fill}}>{\raggedleft\arraybackslash}p{0.19\linewidth}>{\raggedleft\arraybackslash}p{0.20\linewidth}>{\raggedleft\arraybackslash}p{0.26\linewidth}>{\raggedleft\arraybackslash}p{0.20\linewidth}@{}}
            \toprule
            \rowcolor{black!10}
            $s$ & \textit{Grasp} & \textit{Half-cheetah} & \textit{Walker} \\
            \midrule
1 & 98.34 & 73.34 & 7.62 \\
2 & 97.17 & 82.91 & 67.48 \\
3 & 96.68 & 87.01 & 76.27 \\
4 & 96.78 & 86.23 & 75.29 \\
5 & 96.78 & 86.13 & 74.12 \\
            \bottomrule
        \end{tabular*}
    \end{subtable}\hfill
    \begin{subtable}[t]{0.48\linewidth}
        \centering
        \caption{Fixed-policy delay response ($s=5$).}
        \label{tab:app-predictive-delay-data}
        \fontsize{7.2pt}{8pt}\selectfont
        \setlength{\tabcolsep}{3pt}
        \renewcommand{\arraystretch}{0.92}
        \begin{tabular*}{0.84\linewidth}{@{\extracolsep{\fill}}>{\raggedleft\arraybackslash}p{0.16\linewidth}>{\raggedleft\arraybackslash}p{0.17\linewidth}>{\raggedleft\arraybackslash}p{0.22\linewidth}>{\raggedleft\arraybackslash}p{0.17\linewidth}@{}}
            \toprule
            \rowcolor{black!10}
            $D$ (ms) & \textit{Grasp} & \textit{Half-cheetah} & \textit{Walker} \\
            \midrule
0.00 & 96.78 & 86.13 & 74.12 \\
8.33 & 96.39 & 86.72 & 64.94 \\
16.67 & 69.82 & 85.45 & 59.28 \\
25.00 & 41.11 & 83.98 & 55.76 \\
33.33 & 22.17 & 84.57 & 53.81 \\
41.67 & 23.24 & 83.89 & 40.63 \\
50.00 & 24.61 & 82.42 & 37.11 \\
58.33 & 23.05 & 82.32 & 27.93 \\
66.67 & 19.53 & 81.25 & 28.13 \\
            \bottomrule
        \end{tabular*}
    \end{subtable}
    \caption{Calibration accuracy (\%) for Figure~\ref{fig:predictive-case}(1b)--(1c), with $1{,}024$ trials per condition.}
    \label{tab:app-predictive-calibration}
\end{table}

{
For each task, we obtain the composed response on hardware $H$ as
\begin{equation}
    \widehat{P}_{\mathrm{succ},H}(s)
    =\operatorname{clip}\!\left(
        \mathrm{acc}_s\,g\!\left(\alpha_{\mathrm{Ada}}(s)\right)
        h\!\left(T_{\mathrm{inf},H}(s)\right),0,1
    \right).
    \label{eq:app-predictive-composition}
\end{equation}
For a given sampling count $s$, all hardware platforms share the same quality response $g(\alpha_{\mathrm{Ada}}(s))$, while the delay response is evaluated at each platform's inference latency.
With the calibrated responses fixed, we compare predictions with measured accuracy across hardware platforms (Table~\ref{tab:app-predictive-evaluation}).
\par}

\begin{center}
\begin{minipage}{\linewidth}
    \centering

    \footnotesize
    \setlength{\tabcolsep}{2.7pt}
    \renewcommand{\arraystretch}{1.12}
    \begin{tabular}{@{}lr*{4}{rrr}@{}}
        \toprule
        \rowcolor{black!10}
        Task & $s$ & \multicolumn{3}{c}{RTX 6000 Ada} & \multicolumn{3}{c}{RTX 3090} & \multicolumn{3}{c}{AGX Orin 30W} & \multicolumn{3}{c}{AGX Orin 15W} \\
        \cmidrule(lr){3-5}\cmidrule(lr){6-8}\cmidrule(lr){9-11}\cmidrule(lr){12-14}
         & & Pred. & Meas. & Diff. & Pred. & Meas. & Diff. & Pred. & Meas. & Diff. & Pred. & Meas. & Diff. \\
        \midrule
\textit{Grasp} & 1 & 96.94 & 98.44 & $-1.49$ & 96.85 & 98.63 & $-1.78$ & 94.44 & 98.93 & $-4.48$ & 89.57 & 98.05 & $-8.47$ \\
 & 2 & 96.69 & 97.75 & $-1.06$ & 96.44 & 97.56 & $-1.12$ & 82.17 & 91.60 & $-9.43$ & 45.65 & 51.86 & $-6.21$ \\
 & 3 & 96.28 & 97.07 & $-0.79$ & 95.80 & 97.75 & $-1.95$ & 49.10 & 51.17 & $-2.07$ & 24.90 & 24.51 & $+0.39$ \\
 & 4 & 95.85 & 96.88 & $-1.03$ & 94.74 & 97.85 & $-3.11$ & 27.84 & 27.83 & $+0.01$ & 22.44 & 26.27 & $-3.83$ \\
 & 5 & 94.86 & 96.68 & $-1.82$ & 93.21 & 96.29 & $-3.08$ & 23.28 & 22.66 & $+0.62$ & 22.36 & 23.14 & $-0.78$ \\
\midrule
\textit{Half-cheetah} & 1 & 73.17 & 71.19 & $+1.98$ & 73.14 & 73.34 & $-0.20$ & 72.78 & 70.51 & $+2.27$ & 72.53 & 73.63 & $-1.10$ \\
 & 2 & 83.11 & 85.45 & $-2.33$ & 83.05 & 85.06 & $-2.01$ & 82.26 & 83.40 & $-1.14$ & 81.65 & 80.27 & $+1.37$ \\
 & 3 & 85.72 & 87.60 & $-1.88$ & 85.63 & 85.45 & $+0.18$ & 84.38 & 83.50 & $+0.88$ & 83.57 & 82.23 & $+1.34$ \\
 & 4 & 86.06 & 87.70 & $-1.63$ & 85.92 & 86.52 & $-0.61$ & 84.22 & 82.91 & $+1.31$ & 82.96 & 83.30 & $-0.34$ \\
 & 5 & 86.06 & 86.43 & $-0.36$ & 85.92 & 85.64 & $+0.27$ & 83.80 & 83.40 & $+0.40$ & 82.22 & 82.62 & $-0.40$ \\
\midrule
\textit{Walker} & 1 & 7.51 & 6.64 & $+0.87$ & 7.47 & 7.62 & $-0.15$ & 7.04 & 4.79 & $+2.25$ & 6.74 & 3.32 & $+3.42$ \\
 & 2 & 65.73 & 64.06 & $+1.66$ & 65.09 & 64.16 & $+0.93$ & 57.73 & 51.07 & $+6.65$ & 52.00 & 45.80 & $+6.20$ \\
 & 3 & 72.12 & 74.02 & $-1.90$ & 71.21 & 70.51 & $+0.71$ & 58.53 & 57.42 & $+1.11$ & 50.37 & 51.46 & $-1.09$ \\
 & 4 & 71.34 & 71.00 & $+0.35$ & 69.87 & 66.70 & $+3.17$ & 52.80 & 52.64 & $+0.17$ & 40.17 & 36.52 & $+3.65$ \\
 & 5 & 70.00 & 65.14 & $+4.86$ & 68.55 & 64.45 & $+4.10$ & 47.33 & 53.52 & $-6.19$ & 31.45 & 28.81 & $+2.65$ \\
        \bottomrule
    \end{tabular}
    \captionsetup{hypcap=false}
    \captionof{table}{Predicted and measured accuracy (\%) for Figure~\ref{fig:predictive-case}(2). Diff. = Pred. $-$ Meas. (pp).}
    \label{tab:app-predictive-evaluation}
\end{minipage}
\end{center}

\paragraph{DOM.}
The Dynamic Object Manipulation (DOM) benchmark is introduced with DynamicVLA~\cite{xie2026dynamicvla} to evaluate policies on moving-object manipulation,
where the environment continues to evolve during inference delay.
It organizes dynamic scenarios along three dimensions: interaction, perception, and generalization.
We focus on Closed-loop Reactivity (CR), a sub-dimension of interaction that evaluates how quickly a policy adjusts to objects moving at different speeds.
We vary the action expert's sampling steps in $\{1,2,5,10\}$ and inject simulator delays calibrated to emulate inference latency on different hardware platforms.
We evaluate object-motion speed scales of 100\% and 125\% relative to the original dataset.
For each combination of object speed, sampling-step setting, and latency condition, we run 50 trials and report the task success rate and action delay.

\subsection{Result Visualization}
\label{app:simulation-results}

\begin{figure}[t]
    \centering
    \includegraphics[width=\linewidth]{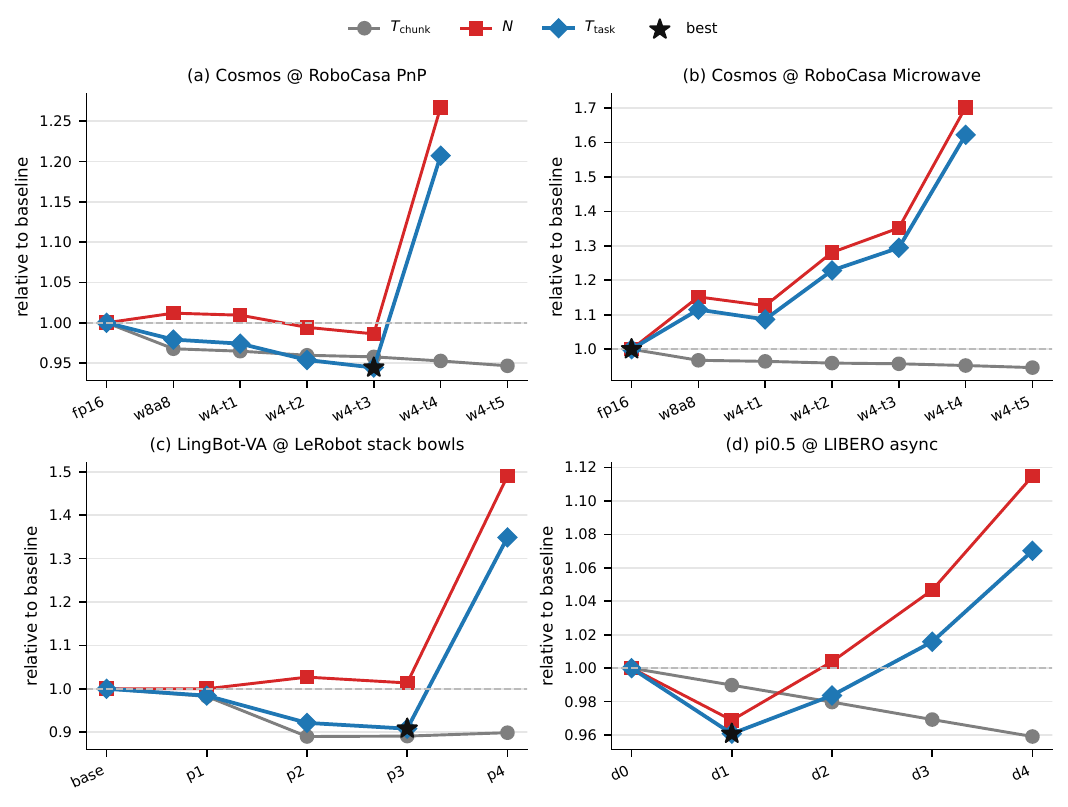}
    \caption{Visualization of the static-task measurements reported in Section 4.2. Values are normalized within each panel by the baseline configuration. In (a), (c), and (d), mild optimization reduces $T_{\mathrm{task}}$, while more aggressive settings eventually increase the required chunk count $N$ enough to reverse the gain, producing an interior sweet spot. In (b), the baseline is already the best measured point, showing that such a sweet spot is not guaranteed and its existence and location are task-dependent.}
    \label{fig:app-obs1-trend}
\end{figure}

Figure~\ref{fig:app-obs1-trend} replots the static-task measurements in Figure~\ref{fig:obs1}(a)--(d) as line charts.
All three quantities are normalized by the corresponding baseline value in each panel.
The star marks the operating point with the lowest $T_{\mathrm{task}}$.

The trends differ across tasks. For example, for the pick-and-place sink-to-counter task in Figure~\ref{fig:app-obs1-trend}(a), $N$ remains roughly stable through moderate quantization, so the latency reduction dominates and the shortest per-task time appears at \texttt{w4-t3}. For the turn-off-microwave task in Figure~\ref{fig:app-obs1-trend}(b), $N$ increases quickly even under mild quantization, so the full-precision baseline still yields the shortest end-to-end time.

\section{Real-World Experiment Details (Section~\ref{sec:real-world})}
\label{app:realworld}

We use a UR5e robot arm~\cite{universalrobotsESeriesRobots} to validate for two task regimes in the physical world. The real-world robot setup follows UMI~\cite{chi2024universalmanipulationinterfaceinthewild}, with the mirrors on the gripper removed. The setup consists of a UR5e robot arm, a UMI gripper, and a GoPro camera for RGB observation. All experiments were conducted with the same robot arm speed setting of 30\% and a control frequency of 10 Hz.

To evaluate how the hardware-dependent inference-to-execution ratio $\rho$ affects $T_{\rm task}$ or task accuracy, we first profile the policy model's inference latency $T_{\rm inf}$ across different hardware platforms. We then introduce a hardware-latency scaling factor $\lambda$ on the
RTX 4070 platform, i.e., $T_{\rm inf,simulated} = \lambda T_{\rm inf,original}, \rho_{\rm simulated} = \lambda \rho_{\rm original}$, which allows us to simulate the inference behavior of different compute platforms under the same task environment.

\subsection{Static Experiment Settings}
\label{app:realworld-static}

We collected 198 expert demonstrations using the UMI gripper and fine-tuned the $\pi_{0.5}$~\cite{black2025pi0.5} model for 30,000 training steps using LoRA~\cite{hu2021loralowrankadaptationlarge}. During LoRA fine-tuning, we initialized the policy from the $\pi_{0.5}$-base model, which consists of two components: a VLM backbone and an action expert. The VLM backbone is PaliGemma~\cite{beyer2024paligemma}, composed of a SigLIP vision encoder and a Gemma-2B language model, while the action expert is a 300M Gemma model. We fine-tuned both components with a learning rate of $2.5 \times 10^{-5}$ and a batch size of 32.

The static manipulation task requires the robot to pick up a red cube on the tabletop and place it into a target box. A trial is considered successful only if the gripper fingers open and release the cube into the target box. Each rollout started from the same predefined initial position.
For each rollout setting, we conducted 5 trials and reported the corresponding task performance.

\subsection{Dynamic Experiment Settings}
\label{app:realworld-dynamic}

We collected 196 expert demonstrations using the UMI gripper and trained the diffusion policy~\cite{chi2024diffusionpolicyvisuomotorpolicy} for 130 epochs. The diffusion policy model is composed of a pretrained ViT vision encoder and a DiT action decoder. The DiT module was trained from scratch with a learning rate of $3.0 \times 10^{-4}$. For the visual encoder, we initialized the ViT backbone from the CLIP~\cite{radford2021learningtransferablevisualmodels} pretrained ViT-L/14 model~\cite{dosovitskiy2021imageworth16x16words} and trained it with a learning rate of $3.0 \times 10^{-5}$ and a batch size of 32.

The dynamic manipulation task requires the robot to grasp the red cube as it moves on a conveyor. We evaluate grasping performance using a three-level score: 2 indicates that the red cube is successfully grasped and held by both gripper fingers; 1 indicates that only one gripper finger contacts the red cube; and 0 indicates that neither gripper finger reaches the red cube. This score reflects the combined effect of reaction speed and motion quality in the dynamic task. The conveyor moved at a constant speed of 6.25 cm/s during evaluation.
For each rollout setting, we conducted 30 trials. Specifically, we used three fixed initial positions and performed 10 trials for each position.

\section{Additional Discussions}
\label{app:discussion}

\subsection{About ``Successful-Only Trial's Chunks''}
\label{app:chunks}

Following the successful-episode reporting protocol in prior work~\cite{li2026inference}, we report chunk number and completion time only over successful rollouts:
\begin{equation}
    N_{\mathrm{succ}}
    =
    \mathbb{E}[N \mid \mathrm{success}],
    \qquad
    T_{\mathrm{task,succ}}
    =
    \mathbb{E}[T_{\mathrm{task}} \mid \mathrm{success}].
    \label{eq:app-success-only}
\end{equation}

The successful-only convention avoids an artifact of simulator rollouts:
when a task fails, the simulator may continue executing additional chunks and corrective attempts until the rollout horizon is reached. Including timeout trajectories would inflate $N$ and $T_{\rm task}$, so the averages would reflect the simulator termination rule instead of the efficiency of completed executions.

\subsection{Stability and Motion Smoothness}
\label{app:smoothness}
\label{app:stability}

\begingroup
\setlength{\parskip}{2pt}
\setlength{\abovedisplayskip}{3pt plus 1pt minus 1pt}
\setlength{\belowdisplayskip}{3pt plus 1pt minus 1pt}
\setlength{\abovedisplayshortskip}{0pt plus 1pt}
\setlength{\belowdisplayshortskip}{3pt plus 1pt minus 1pt}

Prior work~\cite{li2026inference} analyzed in simulation several embodied-efficiency indicators that deteriorate as the degree of lightweighting increases, such as jerk. We additionally conduct real-world experiments on the UR5e platform.

For an end-effector trajectory $\mathbf{x}_0,\ldots,\mathbf{x}_{K-1}$ sampled at interval $\Delta t$, we quantify motion jerk as
\begin{equation}
  \widehat{J}
  =
  \frac{1}{K-3}
  \sum_{k=0}^{K-4}
  \left\|
  \frac{
  \mathbf{x}_{k+3}-3\mathbf{x}_{k+2}
  +3\mathbf{x}_{k+1}-\mathbf{x}_k
  }{\Delta t^3}
  \right\|_2^2.
  \label{eq:app-jerk}
\end{equation}
We report $\widehat{J}$ over static task trials for each step setting in Table~\ref{tab:real-jerk}. The result shows that stronger lightweighting leads to larger motion jerk in real-world execution. This trend is consistent with the simulation results reported in~\cite{li2026inference},
indicating that aggressive lightweighting can reduce the policy model's computation while degrading physical motion quality.

\par\noindent
\begin{minipage}{\linewidth}
\centering
\footnotesize
\captionsetup{font=footnotesize,skip=2pt,hypcap=false}
\captionof{table}{Real-world jerk under different step settings.}
\label{tab:real-jerk}
\fontsize{8pt}{9pt}\selectfont
\begin{tabular}{c c c c c}
\toprule
\rowcolor{black!10} Steps & $1$ & $5$ & $7$ & $10$ \\
\midrule
$\widehat{J}$ ($\mathrm{m^2/s^6}$) & $0.512$ & $0.334$ & $0.313$ & $0.283$ \\
\bottomrule
\end{tabular}
\end{minipage}
\par
\endgroup

\end{document}